\documentclass{article}

\PassOptionsToPackage{square,sort&compress,numbers}{natbib}

\usepackage[preprint]{neurips_2026}

\usepackage[utf8]{inputenc} %
\usepackage[T1]{fontenc}    %

\usepackage{hyperref}       %
\usepackage{url}            %
\usepackage{booktabs}       %
\usepackage{amsfonts}       %
\usepackage{nicefrac}       %
\usepackage{microtype}      %
\usepackage{xcolor}         %

\usepackage{amsmath}
\usepackage{amssymb}
\usepackage{mathtools}
\usepackage{amsfonts}
\usepackage{bm}
\usepackage{siunitx}
\usepackage{longtable}
\usepackage{pdflscape}
\usepackage[inline]{enumitem}
\usepackage{titletoc}
\usepackage{caption}
\usepackage{wrapfig}
\usepackage{layouts}

\usepackage{newfloat}
\DeclareFloatingEnvironment[name=Listing]{listing}

\usepackage[most]{tcolorbox}
\definecolor{rose}{HTML}{fbf6ff}
\definecolor{beige}{HTML}{faf2e9}

\title{Sparsely gated tiny linear experts}

\author{%
  Simon Schug \\
  Department of Computer Science\\
  Princeton University\\
  \texttt{sschug@princeton.edu} \\
}

\begin{document}

\maketitle

\begin{abstract}

Sparsity allows scaling model parameters without proportionally increasing computational cost.
While mixture of experts (MoE) models are made increasingly sparse, individual experts typically remain large and dense.
Here, we demonstrate that further increasing sparsity by shrinking each expert to consist of a single neuron and selecting a tiny fraction of many available neurons can improve compute efficiency and interpretability.
Counterintuitively, the key to achieving both is removing the nonlinearity typically applied to the experts, resulting in a network of sparsely gated \textit{linear} neurons (\texttt{sgatlin}).
In an isoflop comparison, we find that replacing all transformer feedforward layers with \texttt{sgatlin} improves perplexity in language models across different compute budgets.
At the same time, the sparsity and linearity of the resulting feedforward circuits present new opportunities for model interpretability.
In a small-scale case study, we demonstrate that feedforward circuits in \texttt{sgatlin} can be interpreted without having to train additional replacement models.
We find that they form semantically structured clusters and are causally implicated in factual recall.
Our findings paint a possible path towards compute-efficient and interpretable transformer feedforward layers.\footnote{Code available at \url{https://github.com/smonsays/sparsely-gated-linear}}

\end{abstract}

\vspace{-1.5em}
\begin{center}
\includegraphics[width=\textwidth]{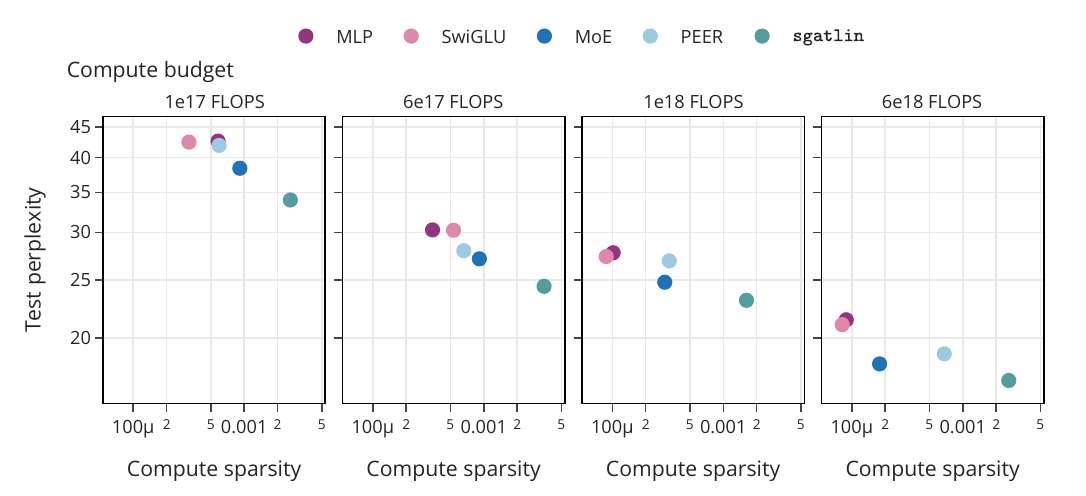}
\end{center}
\vspace{-1em}
\captionof{figure}{\textbf{Sparsity improves compute efficiency.} Comparing feedforward layers in transformer-based language models pretrained on SlimPajama 627B in a compute-optimal setting reveals that increasing compute sparsity -- the ratio of parameters per FLOP -- improves test perplexity across compute budgets. Our sparsely gated linear neuron layer (\texttt{sgatlin}) further increases this ratio.}
\label{fig:graphical-abstract}

\begin{figure}
\includegraphics[width=\textwidth]{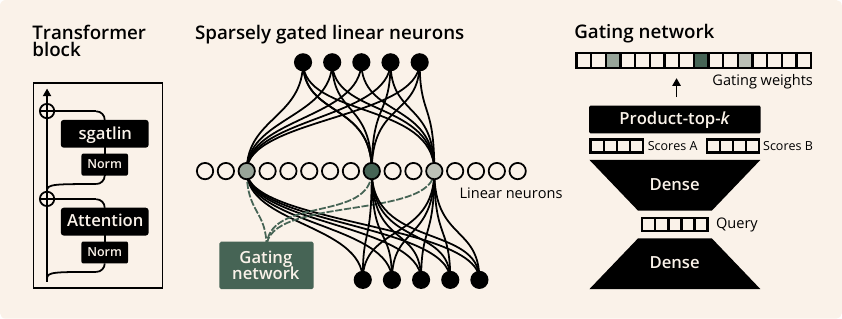}
\caption{\textbf{Sparsely gated linear neurons layer.} \textbf{\textit{Left}} We replace the feedforward layer within each transformer block with a \texttt{sgatlin} layer. \textbf{\textit{Center}} In \texttt{sgatlin}, a gating network selects a sparse subset of $k$ linear neurons whose outputs are linearly combined. \textbf{\textit{Right}} The gating network can efficiently select among a large number of neurons by combining a bottleneck with an efficient top-$k$ operation over the product set of two sets of scores.}
\label{fig:sgatlin-architecture}
\end{figure}

\section{Introduction}

Scale is a key ingredient for creating capable transformer-based models and has played an important part in their success.
At the same time, it demands large amounts of computational resources and yields complex and opaque models that are difficult to understand.
Reconciling scale with compute efficiency and model interpretability is a major challenge.

Feedforward layers in particular consume a considerable amount of the total compute of transformer models as their scale increases.
Since feedforward layers are thought to be the main repository of knowledge, increasing their number of parameters is an important scaling axis.
However, scaling the number of parameters in classical dense feedforward layers leads to a proportional increase in computational cost.
The large number of parameters also presents an obstacle for model interpretability, a challenge compounded by nonlinearities that preclude analysis using standard linear algebra tools.

Sparsity can lower scale-induced computational cost and complexity:
By activating only a subset of parameters, it reduces the compute requirements of parameter scaling and helps localize the subcircuits causally implicated in a given computation \citep{shazeer_outrageously_2017, krajewski_scaling_2024, potapczynski_searching_2024, arora_language_2026}.
However, while mixture of experts (MoE) models are popular and increasingly sparse in current large-scale architectures, they continue to activate billions of parameters per token \citep{muennighoff_olmoe_2024, openai_gpt-oss-120b_2025, google_welcome_2026, qwen_team_qwen3-next_2025, glm-5-team_glm-5_2026, deepseek-ai_deepseek-v4_2026}.

Here, we take a step towards scalable feedforward layers that further realize the potential of compute sparsity by drastically increasing the number of parameters per FLOP.
In our biggest models, less than 0.1\% of feedforward layer parameters are activated per token.
We build on prior work by \citet{he_mixture_2024} that demonstrated how equipping MoEs with an efficient routing mechanism allows scaling parameters at increasing sparsity with fine-grained experts.
By removing the element-wise nonlinearity applied to the experts, eliminating expert parameter sharing across channels, and further decreasing the computational cost of the router, we obtain a layer of sparsely gated linear neurons (\texttt{sgatlin}) that improves language modelling perplexity in a compute-matched setting and opens avenues for model interpretability that do not rely on training additional replacement models.

\vspace{1em}
We summarize our \textbf{main contributions} in the following:

\begin{itemize}[leftmargin=*]
  \item In Section~\ref{sec:sgatlin}, we present \texttt{sgatlin}, a new feedforward layer which sparsely gates a large number of linear neurons and can efficiently scale the number of parameters at sublinear time complexity.
  \item In Section~\ref{sec:language-modelling}, we evaluate the language modelling perplexity of transformers using various feedforward architectures on the SlimPajama 627B dataset \citep{soboleva_slimpajama_2023}. In an isoflop comparison with models of up to 4B parameters, we find that \texttt{sgatlin} performs competitively across compute budgets.
  \item In Section~\ref{sec:model-interpretability}, we conduct an interpretability study, demonstrating that the feedforward circuits of \texttt{sgatlin} form a semantically structured metric space and are causally implicated in factual recall.
\end{itemize}

\section{Sparsely gated linear neurons}
\label{sec:sgatlin}

We now introduce our sparsely gated linear neurons layer (\texttt{sgatlin}), which we will use to replace the feedforward layer in each transformer block (see Figure~\ref{fig:sgatlin-architecture} for an overview).
It consists of two components, a gating network that creates a sparse vector of gating weights, and a large pool of linear neurons -- the experts -- that are linearly combined according to the gating weights.

The gating network computes scores for all available neurons and selects the $k$ largest ones,
\begin{align}
  \label{eq:sgatlin-gating}
  \bm{g} &= \texttt{product-top-k}(\bm{W}^{\text{key}} \bm{W}^{\text{query}} \bm{z}),
\end{align}
where $\bm{W}^{\text{query}} \in \mathbb{R}^{d_{\text{key}} \times d_{\text{model}}}$ and $\bm{W}^{\text{key}} \in \mathbb{R}^{2 \sqrt{d_{\text{ffw}}} \times d_{\text{key}}}$.
The $\texttt{product-top-k}(\bm{x}): \mathbb{R}^{2 \sqrt{d_{\text{ffw}}}} \to \mathbb{R}^{d_{\text{ffw}}}$ operation \citep{lample_large_2019} selects the $k$ largest entries of the product set of two sets of scalars, obtained by slicing the input vector, $\bm{x} \in \mathbb{R}^{2 \sqrt{d_{\text{ffw}}}}$, into two parts of equal size, and returns a sparse vector of gating weights with exactly $k$ non-zero entries, $\bm{g}\in \mathbb{R}^{d_{\text{ffw}}}$ with $\| \bm{g} \|_0 = k$.

The gating weights are then used to linearly combine the output of a sparse subset of $k$ linear neurons,
\begin{align}
  \label{eq:sgatlin-neurons}
  \texttt{sgatlin}(\bm{z}) &= \sum_{i: g_{i} \neq 0} g_{i} \bm{w}_{i}^{\text{out}} \bm{w}_{i}^{\text{in}\top}\bm{z},
\end{align}
where $\bm{w}_{i}^{\text{in}}, \bm{w}_{i}^{\text{out}} \in \mathbb{R}^{d_{\text{model}}}$.
The full \texttt{sgatlin} layer sums over $C$ parallel channels, each of which has distinct sets of parameters except for a shared query projection.
To keep the exposition in the main text concise, we defer the full equations for multiple channels to Appendix~\ref{appsec:sgatlin-channels}.
The resulting layer has several notable properties:

\paragraph{Given the gating weights, the effective circuit processing the layer input is linear.}
Apart from the top-$k$ selection, we do not apply any nonlinearities -- neither to the gating weights nor to the neurons (see Section~\ref{sec:ablations} for an ablation study).
Given the gating weights, the effective circuit processing the layer input is purely linear.
Specifically, each neuron $i$ consists of a rank-one matrix, $\bm{W}_{i}^{\text{neuron}} := \bm{w}_{i}^{\text{out}} \bm{w}_{i}^{\text{in}\top}$, and each token is effectively processed by a (at most) rank-$k$ matrix, $\bm{W}^{\text{circuit}}(\bm{z}) = \sum_{i: g_{i} \neq 0} g_{i} \bm{w}_{i}^{\text{out}} \bm{w}_{i}^{\text{in}\top}$.

\paragraph{Scaling the number of available neurons only incurs a sublinear increase in computational cost.}
Since the effective circuit uses a fixed number of neurons (we use $k=8$ in all experiments), increasing the number of available neurons only affects the time complexity of the gating network.
By keeping the key dimension small and constant (we use $d_{\text{key}}=128$ in all experiments), the time complexity of the gating network scales sublinearly in the number of available neurons, $\mathcal{O}(\sqrt{d_{\text{ffw}}})$.
A comparison of the time and memory complexities of the different feedforward layer variants studied here is shown in Table~\ref{tab:memory-time-complexity-feedforward} in the appendix.

\section{Language modelling}
\label{sec:language-modelling}

In this section, we evaluate decoder-only transformer models in a standard autoregressive language modelling task where we replace the feedforward layers in all transformer blocks.
We will begin by comparing different feedforward layer choices in a compute optimal setting, followed by an ablation study on various architectural components in \texttt{sgatlin}.

\subsection{Isoflop model comparison}

\begin{figure}
  \includegraphics[width=\textwidth]{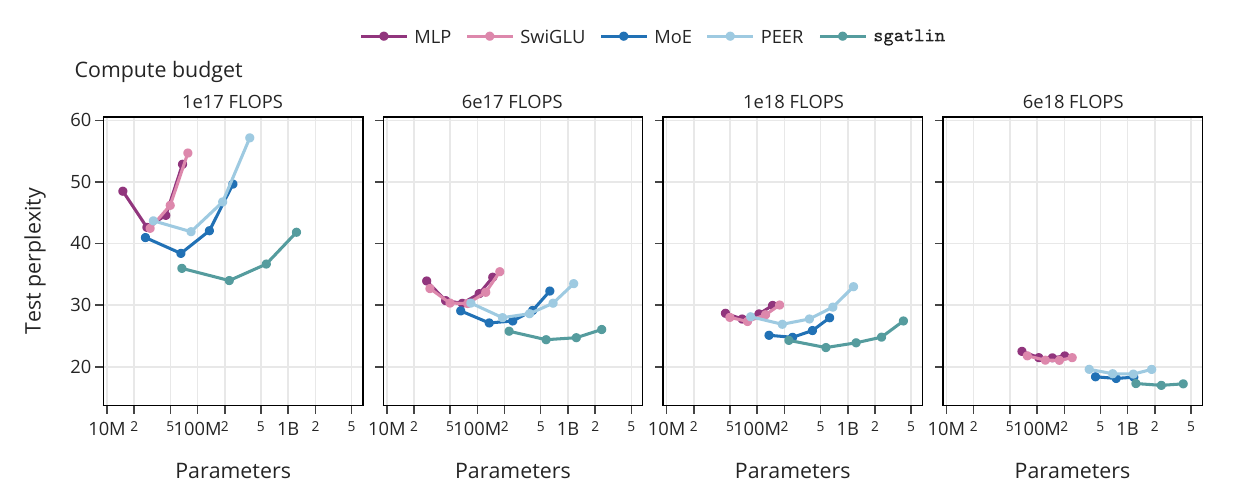}
  \caption{\textbf{Sparsely gated linear neurons improve language modelling performance.} Isoflop comparison of transformer-based language models trained on SlimPajama across four different compute budgets. We compare the choice of transformer feedforward layer type by training models of varying size for a given compute budget and report the resulting perplexities on the test set.}
  \label{fig:isoflop}
\end{figure}

We evaluate how the choice of transformer feedforward layer affects language modelling performance in a compute-matched (isoflop) comparison.
We consider four different compute budgets, $\{ \num{1e17}, \num{6e17}, \num{1e18}, \num{6e18} \}$~FLOPs, for each of which we train models of multiple sizes on the SlimPajama text dataset consisting of 627B tokens \citep{soboleva_slimpajama_2023}.
We compare \texttt{sgatlin} to both dense feedforward layers, a standard GeLU MLP \citep{hendrycks_gaussian_2023} and a SwiGLU layer \citep{shazeer_glu_2020}, as well as two types of mixture of experts layers, the coarse-grained MoE layer used in GPT-OSS \citep{openai_gpt-oss-120b_2025} and the very fine-grained PEER MoE layer \citep{he_mixture_2024}.
To systematically vary model size for a given compute budget, we define a scaling ladder that varies both transformer depth (number of layers) and width (number of feedforward neurons, number of experts), which is detailed in Appendix~\ref{appsec:scaling-ladder}.
We can then determine the number of training tokens that satisfy a particular compute budget based on the model type and its size.

Throughout this section, we consider a context window size of $\num{2048}$ tokens and a batch size of $128$ sequences, resulting in $\num{262144}$ tokens per batch.
We use causally masked multi-head attention in each transformer block, where we set the size of each attention head to $64$ and choose the number of heads according to the model dimension, which is scaled as a multiple of $128$.
We optimize the cross-entropy loss using the AdamW optimizer \citep{loshchilov_decoupled_2019} with a warmup-stable-decay learning rate scheduler \citep{hu_minicpm_2024,hagele_scaling_2024, wen_understanding_2025} that warms up the learning rate from zero for $1000$ steps, followed by a stable learning rate of $\num{1e-3}$ and a final square-root decay over the last 20\% of training tokens back to zero.
The full set of hyperparameters and the procedure for selecting them are provided in Section~\ref{appsec:training-details} of the appendix.

Figure~\ref{fig:isoflop} compares the test perplexities of the different feedforward layer types after training on a given compute budget and model size.
Consistent with prior work \citep[e.g.,][]{fedus_switch_2022, artetxe_efficient_2022, muennighoff_olmoe_2024, lin_moma_2024}, we find that MoE-based models are more compute optimal than dense architectures.
MoE-based models have higher degrees of compute sparsity, allocating more parameters per FLOP, and correspondingly larger model sizes tend to be optimal for a given compute budget.
We plot the best performing model size per compute budget and feedforward type in Figure~\ref{fig:graphical-abstract}, illustrating the relationship between compute sparsity and final test perplexity.
\texttt{sgatlin} has a particularly high compute sparsity and consistently achieves a competitive test perplexity across compute budgets.

\subsection{Ablation study}
\label{sec:ablations}

\begin{wraptable}{R}{0.45\textwidth}
\vspace{-1em}
\caption{\textbf{Model ablations.} Neither adding a nonlinearity to the neurons nor replacing the gating network with the router from PEER \citep{he_mixture_2024} improves language modelling performance of \texttt{sgatlin}.}
\label{tab:ablation}
\begin{tabular}{llr}
\toprule
& \textbf{Ablation} & \textbf{Test perplexity} \\
\midrule
$+$ & ReLU activation & 21.0880 \\
$+$ & GeLU activation & 20.8445 \\
$+$ & Swish activation & 20.6424 \\ \midrule
$\leftrightarrow$ & PEER router & 18.2964 \\ \midrule
$\varnothing$ & \texttt{sgatlin} & 18.1910 \\
\bottomrule
\end{tabular}
\end{wraptable}

The absence of a nonlinear activation function in \texttt{sgatlin} might appear counterintuitive given that it is a standard component in transformer feedforward layers.
Given that MoE layers typically employ an elementwise nonlinear activation function to the experts, we run an ablation study to verify how adding such an activation function in \texttt{sgatlin} affects language modelling performance on our highest compute budget of $\num{6e18}$~FLOPs using $1.2$B-parameter models.
In Table~\ref{tab:ablation}, we report the test perplexity of adding either a ReLU, GeLU, or Swish activation function to \texttt{sgatlin} and find that each choice leads to a decrease in modelling performance.
These findings suggest that the top-$k$ gating by itself suffices to enable nonlinear processing in \texttt{sgatlin}.
We additionally ablate our gating network and swap it with the router used in PEER, confirming that despite our gating networks' reduced time and memory complexity, it leads to comparable performance.

\section{Interpretability}
\label{sec:model-interpretability}

\begin{figure}
  \includegraphics[width=\textwidth]{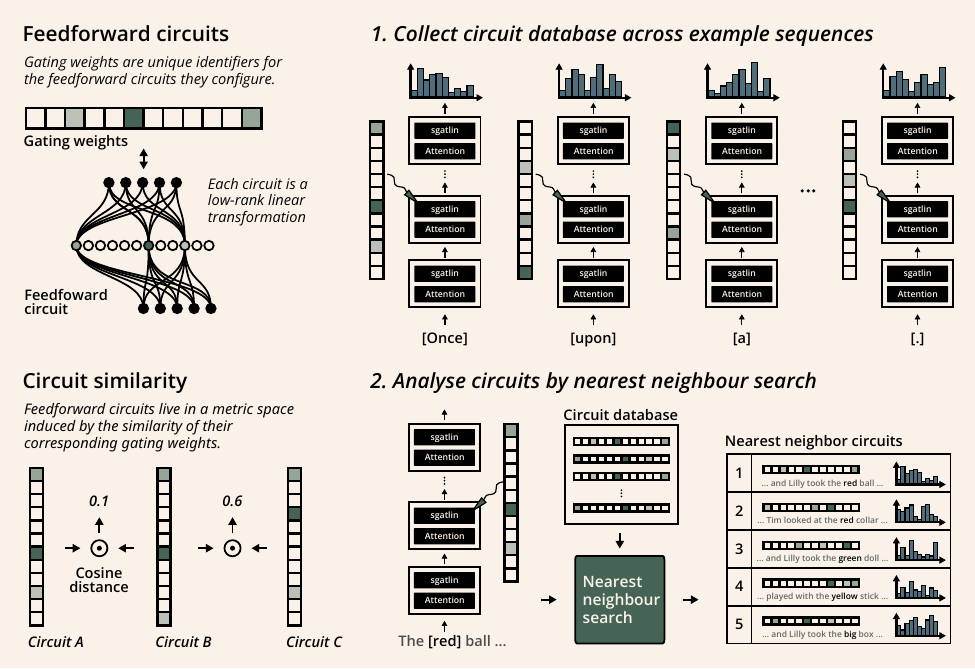}
  \caption{\textbf{Feedforward circuit analysis.} The effective feedforward circuits that process information at a given position in \texttt{sgatlin} are uniquely determined by their gating weights. As a result, the similarity between two circuits can be calculated from the distance between their gating weights. We use this property to analyse a given feedforward circuit by comparing it to similar circuits. For this purpose, we collect a circuit database over a set of reference sequences and retrieve the nearest neighbours from this database to interpret the role of a new circuit.}
  \vspace{-1em}
  \label{fig:circuits-conceptual}
\end{figure}

The structure of \texttt{sgatlin} has interesting implications for model interpretability that we will explore in the following section.
Central to our analysis is the observation that \texttt{sgatlin} linearly combines a small number of \textit{linear} neurons, effectively applying a low-rank linear circuit for each position in the sequence.
Do these feedforward circuits share reusable and possibly interpretable structure?
Instrumental to answering this question is the insight that each circuit is uniquely identified by the sparse gating weights that created it.
We can therefore analyse the set of feedforward circuits in terms of the structure of the metric space of the corresponding gating weights.

To keep this analysis tractable, we will consider a small, \texttt{sgatlin}-based language model trained on the TinyStories corpus \citep{eldan_tinystories_2023}, a collection of short stories for small children which use simple language.
Despite the simplicity of this dataset, models trained on it have been shown to produce coherent and diverse stories with proper grammatical and semantic structure \citep{eldan_tinystories_2023}.
We use the same training pipeline as for our larger models above, albeit with a small, custom tokenizer with a vocabulary size of $8192$, to train a language model with four layers and a context window size of $1024$ tokens.
The full training details can be found in Appendix~\ref{appsec:hyperparameters-tinystories}.

\subsection{Neuron subpopulations form reusable feedforward circuits}

\begin{figure}
\centering
  \includegraphics[width=\textwidth]{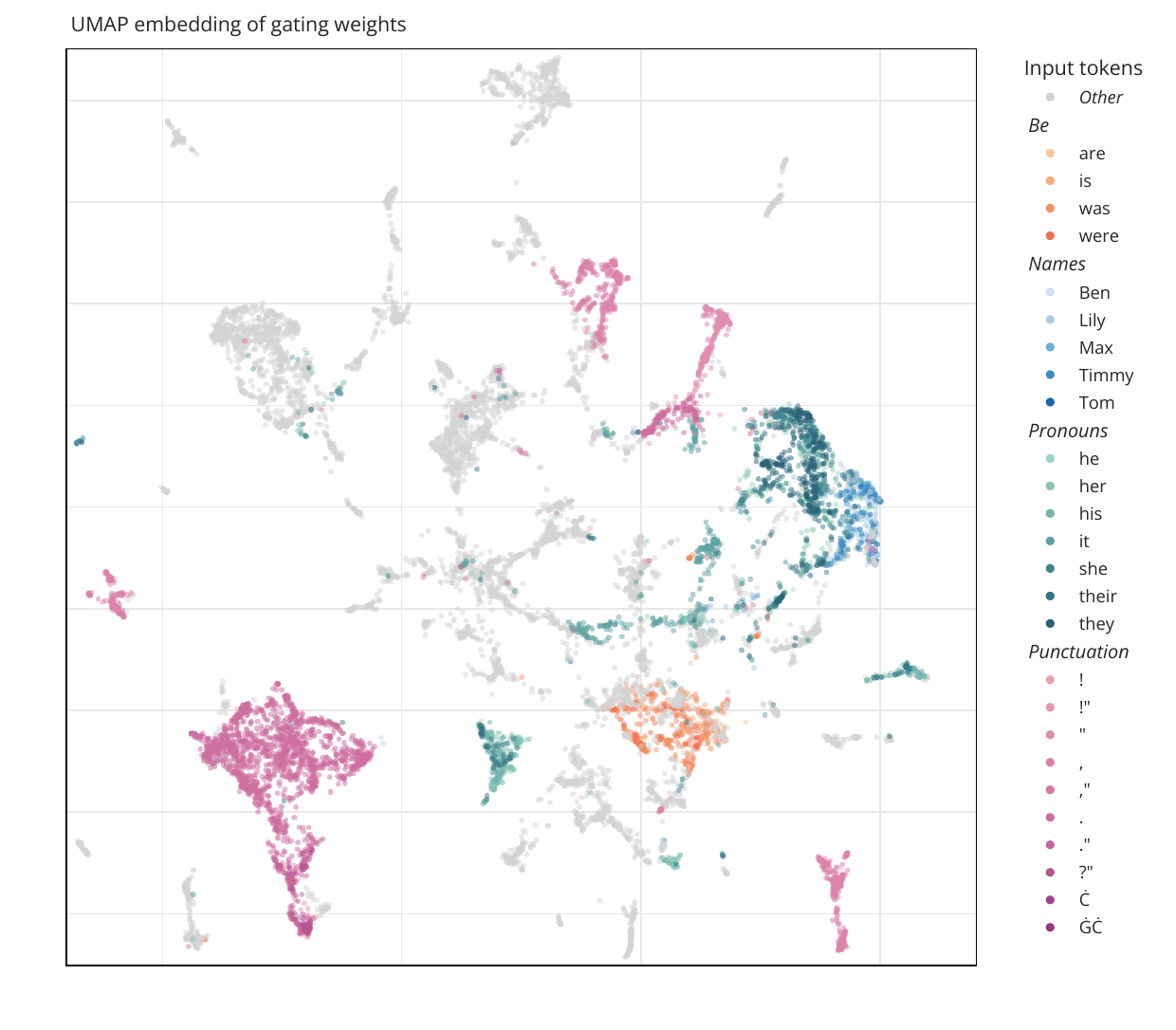}
  \vspace{-2em}
  \caption{\textbf{Feedforward circuits form semantic clusters.} We collect the gating weights of \texttt{sgatlin} of the second layer of a language model trained on TinyStories for the 100 most frequent tokens and embed them into 2D space using UMAP. The resulting embedding forms clusters that are partially explained by the semantics of the corresponding input tokens. For instance, names used in the stories cluster together and pronouns form clusters in their vicinity. An extended version of this figure showing the remaining layers can be found in Figure~\ref{fig:gating-embedding-all-layers}.}
  \label{fig:gating-embedding}
  \vspace{-1em}
\end{figure}

We begin by exploring whether the feedforward circuits activated across novel sequences are reused and form semantically meaningful structure.
For this purpose, we collect the gating weights at each layer as well as the corresponding input tokens for a random subset of $128$ unseen stories from the TinyStories validation set.
Each gating weight is a $d_{\text{ffw}}$-dimensional vector with $k$ non-zero entries.
To visualize the collection of gating weights, we embed them into a two-dimensional space using UMAP \citep{sainburg_parametric_2021} equipped with the cosine distance.
The result of this procedure is shown in Figure~\ref{fig:gating-embedding} for the second feedforward layer, filtered by the $100$ most frequent input tokens in the random subset of unseen stories.
A corresponding plot for all four feedforward layers of the model is shown in Figure~\ref{fig:gating-embedding-all-layers} in the appendix.
Intriguingly, the resulting embeddings have discernible structure with distinct clusters of varying size.
Some -- but not all -- of the structure is related to the input tokens that activated the corresponding circuits.
To illustrate this point, we partially colour points in the two-dimensional embedding according to the input tokens.
The colours are chosen to group semantically related groups of input tokens.
This reveals, for instance, that the larger clusters in the outer perimeter of the plot are made up of circuits that activate for different punctuation tokens.
The different tenses of the verb "to be" are similarly grouped together.
Particularly interesting is that names of the characters in the stories group together in the vicinity of pronouns.
If we track these groups across the different layers in Figure~\ref{fig:gating-embedding-all-layers}, we find that after the second layer, pronouns and names are processed with similar circuits, suggesting that the corresponding circuits operate on the semantic role of these tokens.
Overall, these findings provide some indication that the gating network systematically combines neuron subpopulations into reusable circuits.
This motivates our next analysis.

\subsection{Neighbouring feedforward circuits perform semantically related functions}
\begin{figure}
  \centering
  \includegraphics[width=\textwidth]{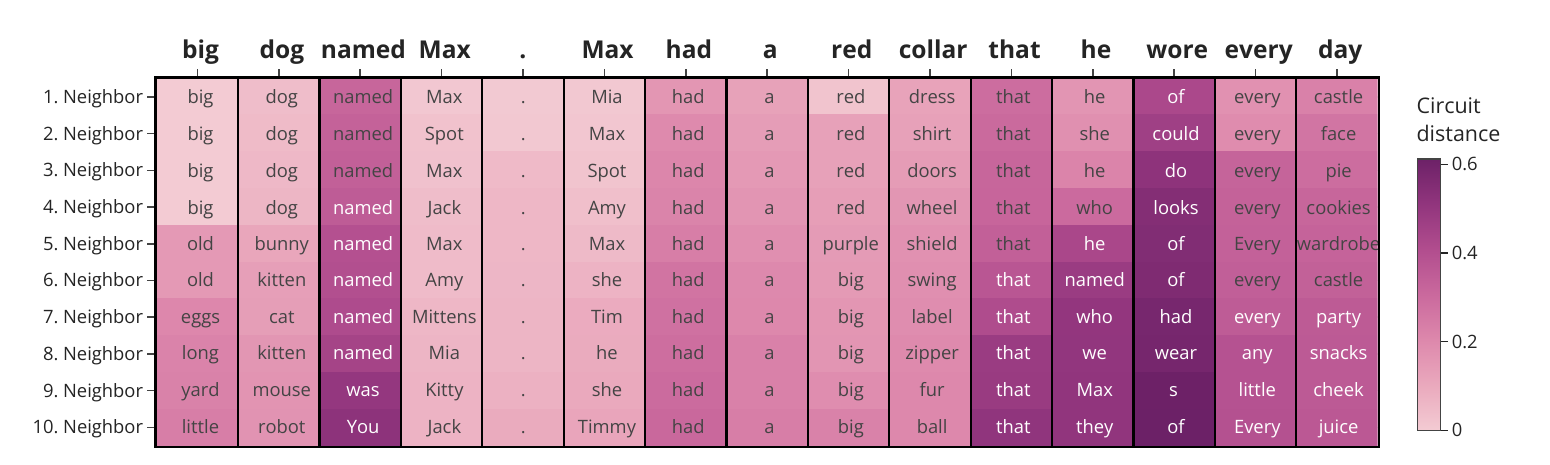}
  \caption{\textbf{Feedforward circuit neighbours.} Following the procedure shown in Figure~\ref{fig:circuits-conceptual} of collecting reference gating weights across input sequences, we can analyse the feedforward circuits for a new input sequence by comparing them to their nearest neighbours among the reference set. In this example, we analyse the feedforward circuits at the penultimate layer for the new input sentence excerpt "\texttt{... big dog named Max. Max had a red collar that he wore every day}". Each column below a respective input token shows the nearest neighbours' input tokens. The colour encodes the distance of a particular neighbour's gating weight to the gating weight of the corresponding input token. For example, the token "he" in our new sentence excerpt has a close neighbouring gating weight that also activated for the input token "he", followed by a neighbour that activated for the input token "she". We show the context input tokens as well as the top output predictions for all neighbours in Table~\ref{tab:gating-neighbors-extended}.}
  \vspace{-1em}
  \label{fig:gating-neighbors}
\end{figure}

Next, we investigate the extent to which a particular feedforward circuit can be understood by relating it to similar circuits.
The overall procedure is outlined in Figure~\ref{fig:circuits-conceptual}.
Once again, we exploit the fact that the gating weights corresponding to each circuit form a metric space.
This allows us to measure the similarity of two circuits by computing the cosine similarity between their corresponding gating weights.
We collect the gating weights corresponding to each input token from unseen sequences in the TinyStories validation set and store them in a database alongside the corresponding input sequence and top output token predictions.
For a novel sequence and its feedforward circuits, we can then search for the closest circuits in the database.
Through this procedure, we aim to establish whether similar circuits have been observed before and whether the circuit is activated by semantically related input tokens or elicits related output predictions.

Figure~\ref{fig:gating-neighbors} shows the input tokens of the ten nearest neighbour circuits of the penultimate layer, along with their distance for a reference sequence.
The extended Table~\ref{tab:gating-neighbors-extended} in the appendix additionally shows input excerpts and output predictions.
We deliberately choose the penultimate feedforward layer here, where we expect the model to perform more abstract computations that go beyond the immediate input or output features.
Some of the reference tokens elicit highly stereotyped circuits that simply activate for exactly the same token across all neighbours.
Intriguingly, however, for several of the tokens in the reference sequence, neighbour circuits activate in different but semantically similar contexts.
For instance, among the neighbours of the feedforward circuit activated for the dog token are circuits that activate for other animals.
Consistent with what we observed in the feedforward circuit embedding in Figure~\ref{fig:gating-embedding}, the feedforward circuit activating for the name "Max" is very similar to circuits that activate for other names.
For some reference tokens, like the "wore" token, the closest neighbours found in the database have a comparably high distance and accordingly bear no apparent semantic relation.

\subsection{Feedforward circuits are causally implicated in factual recall}
\begin{wrapfigure}{R}{0.5\textwidth}
  \centering
  \vspace{-2em}
  \includegraphics[width=\linewidth]{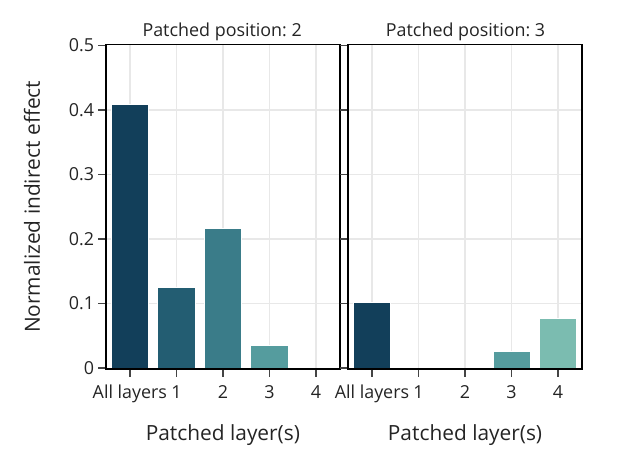}
  \caption{\textbf{Causal intervention on feedforward circuits.} Normalized indirect effect of patching layers at different depth and position in the sequence on counterfactual knowledge pairs extracted from the TinyStories dataset.}
  \label{fig:intervention}
  \vspace{-1em}
\end{wrapfigure}

Finally, we examine how causal interventions on the feedforward circuits impact model predictions.
The fact that \texttt{sgatlin} explicitly separates the circuit formation from its application gives us the chance to delineate a circuit's function from the intermediate data it produces.
Generally speaking, the feedforward layers in transformers are believed to be the primary store of a model's knowledge about the world.
We can use feedforward circuit interventions to assess where such factual knowledge is represented in the feedforward circuits.

To ensure the knowledge we probe for exists in TinyStories and we can reasonably expect the model to have learned it, we extract frequently co-occurring noun-adjective pairs that match the syntactic template "\texttt{The [noun] was [adjective]}" (e.g., "\textit{The turtle was slow}") from the TinyStories training set.
We select the largest set of noun-adjective pairs that ensures no two nouns have the same target adjective.
For the causal intervention, we then create pairs of prompts "\texttt{The [noun] was}" to serve as clean and counterfactual inputs ($x_{\text{clean}}, x_{\text{patch}}$) with divergent nouns and target adjectives.
We can intervene on the forward pass of the clean input by patching the feedforward circuit with the gating weights from the counterfactual input at different token positions and layers in the model.

We quantify the influence of a feedforward circuit, identified by the gating weights $\bm{g}$, with respect to a pair of input sequences, $x_{\text{clean}}$ and $x_{\text{patch}}$, using the normalized indirect effect \citep{pearl_direct_2001}.
Given the corresponding clean and counterfactual target tokens, $y_{\text{clean}}$ and $y_{\text{patch}}$, we measure the logit difference for an input $x$ as  $m(x) = \text{logit}(y_{\text{clean}} \mid x) - \text{logit}(y_{\text{patch}} \mid x)$.
The normalized indirect effect can then be computed as
\begin{align*}
\text{NIE}(m; \bm{g}; x_{\text{clean}}, x_{\text{patch}}) = \frac{m(x_{\text{clean}} \mid \text{do}(\bm{g} = \bm{g}_{\text{patch}})) - m(x_{\text{clean}})}{m(x_{\text{patch}}) - m(x_{\text{clean}})},
\end{align*}
where $\text{do}(\bm{g} = \bm{g}_{\text{patch}})$ denotes the do-intervention on the underlying computational graph, corresponding to clamping the gating weights at specific layers and token positions to the counterfactual gating weights.

We report the normalized indirect effect on the logit difference between the clean and counterfactual adjectives in Figure~\ref{fig:intervention}.
The overall effects we observe are only of moderate magnitude, potentially reflecting that the simple factoids probed for are not only encoded in the feedforward layers but also in other parts of the network such as the input/output embeddings.
Nevertheless, we find that patching the feedforward circuits at the position of the noun has a stronger impact on the normalized indirect effect than patching at the final token prior to the prediction, suggesting that associated factoids are retrieved right when the noun is encountered.
When patching individual layers, the effect is strongest for the second layer of the network, although the outsized effect of patching all layers indicates that feedforward circuits across layers work in conjunction to retrieve a particular factoid.

\section{Related work}

\subsection{Feedforward layers}

The feedforward network architectures we have considered here can be categorized as dense, where all parameters are activated in each forward pass, or mixture of experts (MoE), where only a subset of parameters are active per token.
Interestingly, \texttt{sgatlin} can be conceptually related to both.

The original transformer uses dense ReLU multilayer perceptrons \citep{vaswani_attention_2023}, which implicitly gate the subset of non-negative neurons.
While many transformers replace the ReLU activation with a GeLU due to better modelling performance \citep{radford_improving_2018}, the activation sparsity implicitly induced by ReLU might be useful to improve inference efficiency \citep{mirzadeh_relu_2023}.
We can think of \texttt{sgatlin} as a multilayer perceptron with a special kind of input-dependent nonlinearity that guarantees a fixed activation sparsity.
A similar comparison can be drawn to the dense SwiGLU feedforward layer \citep{shazeer_glu_2020}, in which a second linear input transformation with an ensuing nonlinearity (e.g. sigmoid) is used to softly gate the intermediate neurons via a component-wise product \citep{dauphin_language_2017}.
In a similar vein, \texttt{sgatlin} can be interpreted as hard gating via a component-wise binary mask of fixed sparsity.

As outlined in Section~\ref{sec:sgatlin}, \texttt{sgatlin} can be equally motivated from the perspective of mixture of experts (MoE) layers that select a subset of experts, each of which is a full feedforward network.
Classically, such MoE layers were coarse-grained with large experts, replicating dense layers \citep{fedus_switch_2022}.
More recently, there has been a trend towards fine-grained MoEs with many, smaller experts \citep{krajewski_scaling_2024}.
PEER takes an extreme stance in terms of granularity, making experts as small as possible, with each expert consisting of a single neuron \citep{he_mixture_2024}.
\texttt{sgatlin} is closely related to PEER, albeit with several important architectural changes: In contrast to PEER, \texttt{sgatlin} uses distinct expert parameters for each channel and removes the activation function applied to the experts as well as the softmax nonlinearity applied to the gating scores.

\subsection{Interpretability}

Understanding the internal mechanisms of large-scale transformer-based models is a pressing area of active research given their widespread deployment.
The prevailing approach attempts to automatically extract sparse, interpretable circuits that faithfully characterize a particular behaviour \citep{wang_interpretability_2022, marks_sparse_2024, ameisen_circuit_2025}.
Typically, this involves training replacement models that substitute components of the original model with more interpretable counterparts.
Replacement models such as sparse autoencoders, transcoders, or crosscoders emulate the original model using high-dimensional and sparse representations \citep{cunningham_sparse_2023, dunefsky_transcoders_2024, lindsey_sparse_2024}.
Their shared goal is to decompose the likely polysemantic features of the original model into more easily interpretable, monosemantic features \citep{bricken_towards_2023}.
However, in practice, training such models is costly, surrogate features often only imperfectly approximate the original model \citep{jacovi_towards_2020, gurnee_sae_2024, engels_decomposing_2024}, and sparsity constraints can be hard to reconcile with hierarchical structure, leading to uninterpretable features \citep{bricken_towards_2023, chanin_is_2025}.

An alternative line of work therefore attempts to identify interpretable structure directly in the model under investigation without relying on a replacement model \citep{gao_towards_2026}.
\citet{arora_language_2026} demonstrate that gradient-based attribution can in several instances successfully extract sparse circuits directly from dense feedforward layers.
\citet{herbst_expert_2026} show that as MoEs become more sparsely activated, expert neurons tend to become less polysemantic.
These findings suggest that it may be possible to modify the model architecture itself in order to improve its interpretability.
While the idea of modifying the architecture to improve interpretability is not new, and several attempts have been made to this extent \citep[e.g.,][]{elhage_softmax_2022,sharkey_technical_2023}, such modifications typically lead to a loss in modelling performance or scalability and thus struggle to be adopted.

\section{Discussion}

We have introduced \texttt{sgatlin}, a new feedforward architecture that can serve as a simple drop-in replacement for the feedforward layer in transformers to improve compute efficiency through high levels of sparsity.
In an isoflop comparison, we found competitive language modelling performance across scales and validated key architectural decisions in an ablation study.
The sparse and linear feedforward circuits configured by \texttt{sgatlin} offer interesting avenues for model interpretability, which we explored in a small-scale analysis that revealed that neuron subpopulations form reusable feedforward circuits that live in a semantically structured metric space and are causally implicated in factual recall.

Beyond interpreting feedforward circuits, replacing the feedforward layers with \texttt{sgatlin} might enable new approaches for understanding the inner workings of transformers as a whole.
Similar to how the attention mechanism can be decomposed into a key-query circuit that computes the attention pattern and an output-value circuit that applies a linear transformation to the inputs \citep{elhage_mathematical_2021}, \texttt{sgatlin} can be decomposed into the gating circuit that configures a linear feedforward circuit, which in turn processes the inputs.
This means that conditioned on a particular input's fixed gating weights, attention patterns and layer normalization statistics, the full transformer forward pass becomes effectively a single, large affine transformation followed by the final softmax (a log-linear model).

Linear networks more generally and linearized transformer models in particular have served as important analytical devices and advanced deep learning theory, but ultimately have to concede with simplifying assumptions \citep{saxe_exact_2014, saxe_neural_2022, elhage_toy_2022}.
The decomposition of the transformer into circuit formation and circuit application that can be applied both to the attention mechanism and to \texttt{sgatlin} might open new directions for bridging the gap between theory and practice.

\subsubsection*{Limitations}
Here, we focused on compute sparsity of the feedforward layers in transformer networks, leaving other dense components -- like the attention mechanism -- unchanged, despite their equally important role in terms of compute efficiency and interpretability.
Our model comparison is conducted in an isoflop setting with matched FLOP budgets.
While we believe this is a fair principle for the purpose of our study, it does not directly translate to practical wall clock times during training on current hardware accelerators that are highly optimized for compute dense workloads.
Finally, our interpretability study is conducted on a comparably small dataset and model to provide a controlled setting that allowed us to develop novel techniques for analysing our feedforward layers.
An important next step is to validate these techniques on larger models and integrate them with the analysis of other model components such as the attention mechanism.

\subsubsection*{Broader impacts}
This paper suggests a model architecture that has the potential to improve compute efficiency and interpretability of transformers.
While we foresee no immediate negative societal impact, we hope that it may improve our understanding of this widely deployed technology.

\subsubsection*{Acknowledgments}
We are grateful to Xu Owen He for insightful conversations and feedback early on in the project.
In addition, we would like to thank Yassir Akram, Marc Kaufmann, Nicolas Zucchet, and Seijin Kobayashi for valuable discussions.
Simon Schug was supported by Postdoc.Mobility grant \texttt{P500PT\_225369} from the Swiss National Science Foundation.
This work was supported with $\num{10000}$ GPU hours on the Alps supercomputing infrastructure by the Swiss National Supercomputing Centre under project ID \texttt{a118} as part of the Swiss AI Initiative.

\newpage
\bibliography{bibliography}

\begin{thebibliography}{58}
\providecommand{\natexlab}[1]{#1}
\providecommand{\url}[1]{\texttt{#1}}
\expandafter\ifx\csname urlstyle\endcsname\relax
  \providecommand{\doi}[1]{doi: #1}\else
  \providecommand{\doi}{doi: \begingroup \urlstyle{rm}\Url}\fi

\bibitem[Shazeer et~al.(2017)Shazeer, Mirhoseini, Maziarz, Davis, Le, Hinton,
  and Dean]{shazeer_outrageously_2017}
Noam Shazeer, Azalia Mirhoseini, Krzysztof Maziarz, Andy Davis, Quoc Le,
  Geoffrey Hinton, and Jeff Dean.
\newblock Outrageously {Large} {Neural} {Networks}: {The} {Sparsely}-{Gated}
  {Mixture}-of-{Experts} {Layer}, January 2017.
\newblock URL \url{http://arxiv.org/abs/1701.06538}.
\newblock arXiv:1701.06538 [cs].

\bibitem[Krajewski et~al.(2024)Krajewski, Ludziejewski, Adamczewski, Pióro,
  Krutul, Antoniak, Ciebiera, Król, Odrzygóźdź, Sankowski, Cygan, and
  Jaszczur]{krajewski_scaling_2024}
Jakub Krajewski, Jan Ludziejewski, Kamil Adamczewski, Maciej Pióro, Michał
  Krutul, Szymon Antoniak, Kamil Ciebiera, Krystian Król, Tomasz
  Odrzygóźdź, Piotr Sankowski, Marek Cygan, and Sebastian Jaszczur.
\newblock Scaling {Laws} for {Fine}-{Grained} {Mixture} of {Experts}, February
  2024.
\newblock URL \url{http://arxiv.org/abs/2402.07871}.
\newblock arXiv:2402.07871 [cs].

\bibitem[Potapczynski et~al.(2024)Potapczynski, Qiu, Finzi, Ferri, Chen,
  Goldblum, Bruss, Sa, and Wilson]{potapczynski_searching_2024}
Andres Potapczynski, Shikai Qiu, Marc Finzi, Christopher Ferri, Zixi Chen,
  Micah Goldblum, Bayan Bruss, Christopher~De Sa, and Andrew~Gordon Wilson.
\newblock Searching for {Efficient} {Linear} {Layers} over a {Continuous}
  {Space} of {Structured} {Matrices}, October 2024.
\newblock URL \url{http://arxiv.org/abs/2410.02117}.
\newblock arXiv:2410.02117 [cs].

\bibitem[Arora et~al.(2026)Arora, Wu, Steinhardt, and
  Schwettmann]{arora_language_2026}
Aryaman Arora, Zhengxuan Wu, Jacob Steinhardt, and Sarah Schwettmann.
\newblock Language {Model} {Circuits} {Are} {Sparse} in the {Neuron} {Basis},
  January 2026.
\newblock URL \url{http://arxiv.org/abs/2601.22594}.
\newblock arXiv:2601.22594 [cs].

\bibitem[Muennighoff et~al.(2024)Muennighoff, Soldaini, Groeneveld, Lo,
  Morrison, Min, Shi, Walsh, Tafjord, Lambert, Gu, Arora, Bhagia, Schwenk,
  Wadden, Wettig, Hui, Dettmers, Kiela, Farhadi, Smith, Koh, Singh, and
  Hajishirzi]{muennighoff_olmoe_2024}
Niklas Muennighoff, Luca Soldaini, Dirk Groeneveld, Kyle Lo, Jacob Morrison,
  Sewon Min, Weijia Shi, Pete Walsh, Oyvind Tafjord, Nathan Lambert, Yuling Gu,
  Shane Arora, Akshita Bhagia, Dustin Schwenk, David Wadden, Alexander Wettig,
  Binyuan Hui, Tim Dettmers, Douwe Kiela, Ali Farhadi, Noah~A. Smith, Pang~Wei
  Koh, Amanpreet Singh, and Hannaneh Hajishirzi.
\newblock {OLMoE}: {Open} {Mixture}-of-{Experts} {Language} {Models}, September
  2024.
\newblock URL \url{https://arxiv.org/abs/2409.02060v2}.

\bibitem[{OpenAI} et~al.(2025){OpenAI}, Bao, Ahmad, Ai, Altman, Applebaum,
  Arbus, Arora, Bai, Baker, Bao, Barak, Bennett, Bertao, Brett, Brevdo,
  Brockman, Bubeck, Chang, Chen, Chen, Cheung, Clark, Cook, Dukhan, Dvorak,
  Fives, Fomenko, Garipov, Georgiev, Glaese, Gogineni, Goucher, Gross, Guzman,
  Hallman, Hehir, Heidecke, Helyar, Hu, Huet, Huh, Jain, Johnson, Koch, Kofman,
  Kundel, Kwon, Kyrylov, Le, Leclerc, Lennon, Lessans, Lezcano-Casado, Li, Li,
  Lin, Liss, {Lily}, {Liu}, Liu, Lu, Lu, Martinovic, McCallum, McGrath,
  McKinney, McLaughlin, Mei, Mostovoy, Mu, Myles, Neitz, Nichol, Pachocki,
  Paino, Palmie, Pantuliano, Parascandolo, Park, Pathak, Paz, Peran, Pimenov,
  Pokrass, Proehl, Qiu, Raila, Raso, Ren, Richardson, Robinson, Rotsted,
  Salman, Sanjeev, Schwarzer, Sculley, Sikchi, Simon, Singhal, Song, Stuckey,
  Sun, Tillet, Toizer, Tsimpourlas, Vyas, Wallace, Wang, Wang, Watkins, Weil,
  Wendling, Whinnery, Whitney, Wong, Yang, Yang, Yasunaga, Ying, Zaremba, Zhan,
  Zhang, Zhang, Zhang, and Zhao]{openai_gpt-oss-120b_2025}
{OpenAI}, Haiming Bao, Lama Ahmad, Jason Ai, Sam Altman, Andy Applebaum, Edwin
  Arbus, Rahul~K. Arora, Yu~Bai, Bowen Baker, Haiming Bao, Boaz Barak, Ally
  Bennett, Tyler Bertao, Nivedita Brett, Eugene Brevdo, Greg Brockman,
  Sebastien Bubeck, Che Chang, Kai Chen, Mark Chen, Enoch Cheung, Aidan Clark,
  Dan Cook, Marat Dukhan, Casey Dvorak, Kevin Fives, Vlad Fomenko, Timur
  Garipov, Kristian Georgiev, Mia Glaese, Tarun Gogineni, Adam Goucher, Lukas
  Gross, Katia~Gil Guzman, John Hallman, Jackie Hehir, Johannes Heidecke, Alec
  Helyar, Haitang Hu, Romain Huet, Jacob Huh, Saachi Jain, Zach Johnson, Chris
  Koch, Irina Kofman, Dominik Kundel, Jason Kwon, Volodymyr Kyrylov, Elaine~Ya
  Le, Guillaume Leclerc, James~Park Lennon, Scott Lessans, Mario
  Lezcano-Casado, Yuanzhi Li, Zhuohan Li, Ji~Lin, Jordan Liss, {Lily}, {Liu},
  Jiancheng Liu, Kevin Lu, Chris Lu, Zoran Martinovic, Lindsay McCallum, Josh
  McGrath, Scott McKinney, Aidan McLaughlin, Song Mei, Steve Mostovoy, Tong Mu,
  Gideon Myles, Alexander Neitz, Alex Nichol, Jakub Pachocki, Alex Paino, Dana
  Palmie, Ashley Pantuliano, Giambattista Parascandolo, Jongsoo Park, Leher
  Pathak, Carolina Paz, Ludovic Peran, Dmitry Pimenov, Michelle Pokrass,
  Elizabeth Proehl, Huida Qiu, Gaby Raila, Filippo Raso, Hongyu Ren, Kimmy
  Richardson, David Robinson, Bob Rotsted, Hadi Salman, Suvansh Sanjeev, Max
  Schwarzer, D.~Sculley, Harshit Sikchi, Kendal Simon, Karan Singhal, Yang
  Song, Dane Stuckey, Zhiqing Sun, Philippe Tillet, Sam Toizer, Foivos
  Tsimpourlas, Nikhil Vyas, Eric Wallace, Xin Wang, Miles Wang, Olivia Watkins,
  Kevin Weil, Amy Wendling, Kevin Whinnery, Cedric Whitney, Hannah Wong, Lin
  Yang, Yu~Yang, Michihiro Yasunaga, Kristen Ying, Wojciech Zaremba, Wenting
  Zhan, Cyril Zhang, Brian Zhang, Eddie Zhang, and Shengjia Zhao.
\newblock gpt-oss-120b \& gpt-oss-20b {Model} {Card}, 2025.
\newblock URL \url{https://arxiv.org/abs/2508.10925}.
\newblock \_eprint: 2508.10925.

\bibitem[{Google}(2026)]{google_welcome_2026}
{Google}.
\newblock Welcome {Gemma} 4: {Frontier} multimodal intelligence on device,
  April 2026.
\newblock URL \url{https://huggingface.co/blog/gemma4}.

\bibitem[Team(2025)]{qwen_team_qwen3-next_2025}
Qwen Team.
\newblock Qwen3-{Next}: {Towards} {Ultimate} {Training} \& {Inference}
  {Efficiency}, September 2025.
\newblock URL \url{https://huggingface.co/Qwen/Qwen3-Next-80B-A3B-Instruct}.

\bibitem[GLM-5-Team et~al.(2026)GLM-5-Team, Zeng, Lv, Hou, Du, Zheng, Chen,
  Yin, Ge, Huang, Xie, Zhu, Yin, Wang, Pan, Zeng, Zhang, Wang, Chen, Zhang,
  Jiao, Guo, Wang, Du, Wu, Wang, Li, Fan, Zhong, Liu, Zhao, Du, Dong, Lu,
  Shuang-Li, Cao, Liu, Jiang, Chen, Zhang, Huang, Dong, Xu, Wei, An, Niu, Zhu,
  Wen, Cen, Bai, Qiao, Wang, Wang, Zhu, Liu, Li, Wang, Wen, Huang, Cai, Yu, Li,
  Hu, Zhang, Zhang, Lin, Yang, Wang, Ai, Zhu, Yi, Chen, Wen, Sun, Zhao, Hu,
  Zhang, Liu, Zhang, Peng, Tai, Zhang, Liu, Wang, Yan, Ge, Liu, Chu, Zhao,
  Wang, Zhao, Ren, Wang, Zhang, Gui, Zhao, Li, An, Li, Yuan, Du, Liu, Zhi,
  Duan, Zhou, Wei, Wang, Luo, Zhang, Sha, Xu, Wu, Ding, Chen, Li, Lin, Ta, Zou,
  Song, Yang, Tu, Yang, Wu, Zhang, Li, Li, Fan, Qin, Tian, Zhang, Yu, Liang,
  Kuang, Cheng, Li, Yan, Hu, Ling, Fan, Xia, Zhang, Zhang, Pan, Zou, Zhang,
  Liu, Wu, Li, Wang, Zhu, Tan, Zhou, Pan, Zhang, Su, Geng, Yan, Tan, Bi, Shen,
  Yang, Li, Liu, Wang, Li, Wu, Zhang, Duan, Zhang, Liu, Jiang, Yan, Zhang, Wei,
  Chen, Feng, Yao, Chai, Wang, Zhang, Xu, Huang, Wang, Li, Dong, and
  Tang]{glm-5-team_glm-5_2026}
GLM-5-Team, Aohan Zeng, Xin Lv, Zhenyu Hou, Zhengxiao Du, Qinkai Zheng, Bin
  Chen, Da~Yin, Chendi Ge, Chenghua Huang, Chengxing Xie, Chenzheng Zhu,
  Congfeng Yin, Cunxiang Wang, Gengzheng Pan, Hao Zeng, Haoke Zhang, Haoran
  Wang, Huilong Chen, Jiajie Zhang, Jian Jiao, Jiaqi Guo, Jingsen Wang,
  Jingzhao Du, Jinzhu Wu, Kedong Wang, Lei Li, Lin Fan, Lucen Zhong, Mingdao
  Liu, Mingming Zhao, Pengfan Du, Qian Dong, Rui Lu, Shuang-Li, Shulin Cao,
  Song Liu, Ting Jiang, Xiaodong Chen, Xiaohan Zhang, Xuancheng Huang, Xuezhen
  Dong, Yabo Xu, Yao Wei, Yifan An, Yilin Niu, Yitong Zhu, Yuanhao Wen, Yukuo
  Cen, Yushi Bai, Zhongpei Qiao, Zihan Wang, Zikang Wang, Zilin Zhu, Ziqiang
  Liu, Zixuan Li, Bojie Wang, Bosi Wen, Can Huang, Changpeng Cai, Chao Yu, Chen
  Li, Chengwei Hu, Chenhui Zhang, Dan Zhang, Daoyan Lin, Dayong Yang, Di~Wang,
  Ding Ai, Erle Zhu, Fangzhou Yi, Feiyu Chen, Guohong Wen, Hailong Sun, Haisha
  Zhao, Haiyi Hu, Hanchen Zhang, Hanrui Liu, Hanyu Zhang, Hao Peng, Hao Tai,
  Haobo Zhang, He~Liu, Hongwei Wang, Hongxi Yan, Hongyu Ge, Huan Liu, Huanpeng
  Chu, Jia'ni Zhao, Jiachen Wang, Jiajing Zhao, Jiamin Ren, Jiapeng Wang,
  Jiaxin Zhang, Jiayi Gui, Jiayue Zhao, Jijie Li, Jing An, Jing Li, Jingwei
  Yuan, Jinhua Du, Jinxin Liu, Junkai Zhi, Junwen Duan, Kaiyue Zhou, Kangjian
  Wei, Ke~Wang, Keyun Luo, Laiqiang Zhang, Leigang Sha, Liang Xu, Lindong Wu,
  Lintao Ding, Lu~Chen, Minghao Li, Nianyi Lin, Pan Ta, Qiang Zou, Rongjun
  Song, Ruiqi Yang, Shangqing Tu, Shangtong Yang, Shaoxiang Wu, Shengyan Zhang,
  Shijie Li, Shuang Li, Shuyi Fan, Wei Qin, Wei Tian, Weining Zhang, Wenbo Yu,
  Wenjie Liang, Xiang Kuang, Xiangmeng Cheng, Xiangyang Li, Xiaoquan Yan,
  Xiaowei Hu, Xiaoying Ling, Xing Fan, Xingye Xia, Xinyuan Zhang, Xinze Zhang,
  Xirui Pan, Xu~Zou, Xunkai Zhang, Yadi Liu, Yandong Wu, Yanfu Li, Yidong Wang,
  Yifan Zhu, Yijun Tan, Yilin Zhou, Yiming Pan, Ying Zhang, Yinpei Su, Yipeng
  Geng, Yong Yan, Yonglin Tan, Yuean Bi, Yuhan Shen, Yuhao Yang, Yujiang Li,
  Yunan Liu, Yunqing Wang, Yuntao Li, Yurong Wu, Yutao Zhang, Yuxi Duan, Yuxuan
  Zhang, Zezhen Liu, Zhengtao Jiang, Zhenhe Yan, Zheyu Zhang, Zhixiang Wei,
  Zhuo Chen, Zhuoer Feng, Zijun Yao, Ziwei Chai, Ziyuan Wang, Zuzhou Zhang, Bin
  Xu, Minlie Huang, Hongning Wang, Juanzi Li, Yuxiao Dong, and Jie Tang.
\newblock {GLM}-5: from {Vibe} {Coding} to {Agentic} {Engineering}, February
  2026.
\newblock URL \url{http://arxiv.org/abs/2602.15763}.
\newblock arXiv:2602.15763 [cs].

\bibitem[{DeepSeek-AI}(2026)]{deepseek-ai_deepseek-v4_2026}
{DeepSeek-AI}.
\newblock {DeepSeek}-{V4}: {Towards} {Highly} {Efficient} {Million}-{Token}
  {Context} {Intelligence}, 2026.
\newblock URL \url{https://huggingface.co/deepseek-ai/DeepSeek-V4-Flash}.

\bibitem[He(2024)]{he_mixture_2024}
Xu~Owen He.
\newblock Mixture of {A} {Million} {Experts}, July 2024.
\newblock URL \url{http://arxiv.org/abs/2407.04153}.
\newblock arXiv:2407.04153.

\bibitem[Soboleva et~al.(2023)Soboleva, Al-Khateeb, Myers, Steeves, Hestness,
  and Dey]{soboleva_slimpajama_2023}
Daria Soboleva, Faisal Al-Khateeb, Robert Myers, Jacob~R Steeves, Joel
  Hestness, and Nolan Dey.
\newblock {SlimPajama}: {A} {627B} token cleaned and deduplicated version of
  {RedPajama}, June 2023.
\newblock URL \url{https://huggingface.co/datasets/cerebras/SlimPajama-627B}.

\bibitem[Lample et~al.(2019)Lample, Sablayrolles, Ranzato, Denoyer, and
  Jégou]{lample_large_2019}
Guillaume Lample, Alexandre Sablayrolles, Marc'Aurelio Ranzato, Ludovic
  Denoyer, and Hervé Jégou.
\newblock Large {Memory} {Layers} with {Product} {Keys}, December 2019.
\newblock URL \url{http://arxiv.org/abs/1907.05242}.
\newblock arXiv:1907.05242.

\bibitem[Hendrycks and Gimpel(2023)]{hendrycks_gaussian_2023}
Dan Hendrycks and Kevin Gimpel.
\newblock Gaussian {Error} {Linear} {Units} ({GELUs}), 2023.
\newblock URL \url{https://arxiv.org/abs/1606.08415}.
\newblock \_eprint: 1606.08415.

\bibitem[Shazeer(2020)]{shazeer_glu_2020}
Noam Shazeer.
\newblock {GLU} {Variants} {Improve} {Transformer}, 2020.
\newblock URL \url{https://arxiv.org/abs/2002.05202}.
\newblock arXiv: 2002.05202.

\bibitem[Loshchilov and Hutter(2019)]{loshchilov_decoupled_2019}
Ilya Loshchilov and Frank Hutter.
\newblock Decoupled {Weight} {Decay} {Regularization}, January 2019.
\newblock URL \url{http://arxiv.org/abs/1711.05101}.
\newblock arXiv:1711.05101 [cs, math].

\bibitem[Hu et~al.(2024)Hu, Tu, Han, Cui, He, Zhao, Long, Zheng, Fang, Huang,
  Zhang, Thai, Wang, Yao, Zhao, Zhou, Cai, Zhai, Ding, Jia, Zeng, Li, Liu, and
  Sun]{hu_minicpm_2024}
Shengding Hu, Yuge Tu, Xu~Han, Ganqu Cui, Chaoqun He, Weilin Zhao, Xiang Long,
  Zhi Zheng, Yewei Fang, Yuxiang Huang, Xinrong Zhang, Zhen~Leng Thai, Chongyi
  Wang, Yuan Yao, Chenyang Zhao, Jie Zhou, Jie Cai, Zhongwu Zhai, Ning Ding,
  Chao Jia, Guoyang Zeng, Dahai Li, Zhiyuan Liu, and Maosong Sun.
\newblock {MiniCPM}: {Unveiling} the {Potential} of {Small} {Language} {Models}
  with {Scalable} {Training} {Strategies}.
\newblock August 2024.
\newblock URL \url{https://openreview.net/forum?id=3X2L2TFr0f}.

\bibitem[Hägele et~al.(2024)Hägele, Bakouch, Kosson, Allal, Werra, and
  Jaggi]{hagele_scaling_2024}
Alexander Hägele, Elie Bakouch, Atli Kosson, Loubna~Ben Allal, Leandro~Von
  Werra, and Martin Jaggi.
\newblock Scaling {Laws} and {Compute}-{Optimal} {Training} {Beyond} {Fixed}
  {Training} {Durations}.
\newblock November 2024.
\newblock URL \url{https://openreview.net/forum?id=Y13gSfTjGr}.

\bibitem[Wen et~al.(2025)Wen, Li, Wang, Hall, Liang, and
  Ma]{wen_understanding_2025}
Kaiyue Wen, Zhiyuan Li, Jason~S. Wang, David Leo~Wright Hall, Percy Liang, and
  Tengyu Ma.
\newblock Understanding {Warmup}-{Stable}-{Decay} {Learning} {Rates}: {A}
  {River} {Valley} {Loss} {Landscape} {View}.
\newblock In \emph{The {Thirteenth} {International} {Conference} on {Learning}
  {Representations}}, 2025.
\newblock URL \url{https://openreview.net/forum?id=m51BgoqvbP}.

\bibitem[Fedus et~al.(2022)Fedus, Zoph, and Shazeer]{fedus_switch_2022}
William Fedus, Barret Zoph, and Noam Shazeer.
\newblock Switch {Transformers}: {Scaling} to {Trillion} {Parameter} {Models}
  with {Simple} and {Efficient} {Sparsity}.
\newblock \emph{Journal of Machine Learning Research}, 23\penalty0
  (120):\penalty0 1--39, 2022.
\newblock ISSN 1533-7928.
\newblock URL \url{http://jmlr.org/papers/v23/21-0998.html}.

\bibitem[Artetxe et~al.(2022)Artetxe, Bhosale, Goyal, Mihaylov, Ott, Shleifer,
  Lin, Du, Iyer, Pasunuru, Anantharaman, Li, Chen, Akin, Baines, Martin, Zhou,
  Koura, O'Horo, Wang, Zettlemoyer, Diab, Kozareva, and
  Stoyanov]{artetxe_efficient_2022}
Mikel Artetxe, Shruti Bhosale, Naman Goyal, Todor Mihaylov, Myle Ott, Sam
  Shleifer, Xi~Victoria Lin, Jingfei Du, Srinivasan Iyer, Ramakanth Pasunuru,
  Giridharan Anantharaman, Xian Li, Shuohui Chen, Halil Akin, Mandeep Baines,
  Louis Martin, Xing Zhou, Punit~Singh Koura, Brian O'Horo, Jeffrey Wang, Luke
  Zettlemoyer, Mona Diab, Zornitsa Kozareva, and Veselin Stoyanov.
\newblock Efficient {Large} {Scale} {Language} {Modeling} with {Mixtures} of
  {Experts}.
\newblock In Yoav Goldberg, Zornitsa Kozareva, and Yue Zhang, editors,
  \emph{Proceedings of the 2022 {Conference} on {Empirical} {Methods} in
  {Natural} {Language} {Processing}}, pages 11699--11732, Abu Dhabi, United
  Arab Emirates, December 2022. Association for Computational Linguistics.
\newblock \doi{10.18653/v1/2022.emnlp-main.804}.
\newblock URL \url{https://aclanthology.org/2022.emnlp-main.804/}.

\bibitem[Lin et~al.(2024)Lin, Shrivastava, Luo, Iyer, Lewis, Ghosh,
  Zettlemoyer, and Aghajanyan]{lin_moma_2024}
Xi~Victoria Lin, Akshat Shrivastava, Liang Luo, Srinivasan Iyer, Mike Lewis,
  Gargi Ghosh, Luke Zettlemoyer, and Armen Aghajanyan.
\newblock {MoMa}: {Efficient} {Early}-{Fusion} {Pre}-training with {Mixture} of
  {Modality}-{Aware} {Experts}, July 2024.
\newblock URL \url{https://arxiv.org/abs/2407.21770v3}.

\bibitem[Eldan and Li(2023)]{eldan_tinystories_2023}
Ronen Eldan and Yuanzhi Li.
\newblock {TinyStories}: {How} {Small} {Can} {Language} {Models} {Be} and
  {Still} {Speak} {Coherent} {English}?, May 2023.
\newblock URL \url{http://arxiv.org/abs/2305.07759}.
\newblock arXiv:2305.07759 [cs].

\bibitem[Sainburg et~al.(2021)Sainburg, McInnes, and
  Gentner]{sainburg_parametric_2021}
Tim Sainburg, Leland McInnes, and Timothy~Q Gentner.
\newblock Parametric {UMAP} {Embeddings} for {Representation} and
  {Semisupervised} {Learning}.
\newblock \emph{Neural Computation}, 33\penalty0 (11):\penalty0 2881--2907,
  2021.

\bibitem[Pearl(2001)]{pearl_direct_2001}
Judea Pearl.
\newblock Direct and indirect effects.
\newblock In \emph{Proceedings of the {Seventeenth} conference on {Uncertainty}
  in artificial intelligence}, {UAI}'01, pages 411--420, San Francisco, CA,
  USA, August 2001. Morgan Kaufmann Publishers Inc.
\newblock ISBN 978-1-55860-800-9.
\newblock URL \url{https://dl.acm.org/doi/10.5555/2074022.2074073}.

\bibitem[Vaswani et~al.(2023)Vaswani, Shazeer, Parmar, Uszkoreit, Jones, Gomez,
  Kaiser, and Polosukhin]{vaswani_attention_2023}
Ashish Vaswani, Noam Shazeer, Niki Parmar, Jakob Uszkoreit, Llion Jones,
  Aidan~N. Gomez, Lukasz Kaiser, and Illia Polosukhin.
\newblock Attention {Is} {All} {You} {Need}, August 2023.
\newblock URL \url{http://arxiv.org/abs/1706.03762}.
\newblock arXiv:1706.03762 [cs].

\bibitem[Radford et~al.(2018)Radford, Narasimhan, Salimans, Sutskever, and
  {others}]{radford_improving_2018}
Alec Radford, Karthik Narasimhan, Tim Salimans, Ilya Sutskever, and {others}.
\newblock Improving language understanding by generative pre-training, 2018.

\bibitem[Mirzadeh et~al.(2023)Mirzadeh, Alizadeh, Mehta, Mundo, Tuzel, Samei,
  Rastegari, and Farajtabar]{mirzadeh_relu_2023}
Iman Mirzadeh, Keivan Alizadeh, Sachin Mehta, Carlo C.~Del Mundo, Oncel Tuzel,
  Golnoosh Samei, Mohammad Rastegari, and Mehrdad Farajtabar.
\newblock {ReLU} {Strikes} {Back}: {Exploiting} {Activation} {Sparsity} in
  {Large} {Language} {Models}, 2023.
\newblock URL \url{https://arxiv.org/abs/2310.04564}.

\bibitem[Dauphin et~al.(2017)Dauphin, Fan, Auli, and
  Grangier]{dauphin_language_2017}
Yann~N. Dauphin, Angela Fan, Michael Auli, and David Grangier.
\newblock Language {Modeling} with {Gated} {Convolutional} {Networks}.
\newblock In \emph{Proceedings of the 34th {International} {Conference} on
  {Machine} {Learning}}, pages 933--941. PMLR, July 2017.
\newblock URL \url{https://proceedings.mlr.press/v70/dauphin17a.html}.

\bibitem[Wang et~al.(2022)Wang, Variengien, Conmy, Shlegeris, and
  Steinhardt]{wang_interpretability_2022}
Kevin Wang, Alexandre Variengien, Arthur Conmy, Buck Shlegeris, and Jacob
  Steinhardt.
\newblock Interpretability in the {Wild}: a {Circuit} for {Indirect} {Object}
  {Identification} in {GPT}-2 small, November 2022.
\newblock URL \url{http://arxiv.org/abs/2211.00593}.
\newblock arXiv:2211.00593 [cs].

\bibitem[Marks et~al.(2024)Marks, Rager, Michaud, Belinkov, Bau, and
  Mueller]{marks_sparse_2024}
Samuel Marks, Can Rager, Eric~J. Michaud, Yonatan Belinkov, David Bau, and
  Aaron Mueller.
\newblock Sparse {Feature} {Circuits}: {Discovering} and {Editing}
  {Interpretable} {Causal} {Graphs} in {Language} {Models}.
\newblock October 2024.
\newblock URL \url{https://openreview.net/forum?id=I4e82CIDxv}.

\bibitem[Ameisen et~al.(2025)Ameisen, Lindsey, Pearce, Gurnee, Turner, Chen,
  Citro, Abrahams, Carter, Hosmer, Marcus, Sklar, Templeton, Bricken,
  McDougall, Cunningham, Henighan, Jermyn, Jones, Persic, Qi, Thompson,
  Zimmerman, Rivoire, Conerly, Olah, and Batson]{ameisen_circuit_2025}
Emmanuel Ameisen, Jack Lindsey, Adam Pearce, Wes Gurnee, Nicholas~L. Turner,
  Brian Chen, Craig Citro, David Abrahams, Shan Carter, Basil Hosmer, Jonathan
  Marcus, Michael Sklar, Adly Templeton, Trenton Bricken, Callum McDougall,
  Hoagy Cunningham, Thomas Henighan, Adam Jermyn, Andy Jones, Andrew Persic,
  Zhenyi Qi, T.~Ben Thompson, Sam Zimmerman, Kelley Rivoire, Thomas Conerly,
  Chris Olah, and Joshua Batson.
\newblock Circuit {Tracing}: {Revealing} {Computational} {Graphs} in {Language}
  {Models}, 2025.

\bibitem[Cunningham et~al.(2023)Cunningham, Ewart, Riggs, Huben, and
  Sharkey]{cunningham_sparse_2023}
Hoagy Cunningham, Aidan Ewart, Logan Riggs, Robert Huben, and Lee Sharkey.
\newblock Sparse {Autoencoders} {Find} {Highly} {Interpretable} {Features} in
  {Language} {Models}, October 2023.
\newblock URL \url{http://arxiv.org/abs/2309.08600}.
\newblock arXiv:2309.08600 [cs].

\bibitem[Dunefsky et~al.(2024)Dunefsky, Chlenski, and
  Nanda]{dunefsky_transcoders_2024}
Jacob Dunefsky, Philippe Chlenski, and Neel Nanda.
\newblock Transcoders {Find} {Interpretable} {LLM} {Feature} {Circuits},
  November 2024.
\newblock URL \url{http://arxiv.org/abs/2406.11944}.
\newblock arXiv:2406.11944 [cs].

\bibitem[Lindsey et~al.(2024)Lindsey, Templeton, Marcus, Conerly, Batson, and
  Olah]{lindsey_sparse_2024}
J.~Lindsey, A.~Templeton, J.~Marcus, T.~Conerly, J.~Batson, and C.~Olah.
\newblock Sparse {Crosscoders} for {Cross}-{Layer} {Features} and {Model}
  {Diffing}, 2024.
\newblock URL \url{https://transformer-circuits.pub/2024/crosscoders/}.

\bibitem[Bricken et~al.(2023)Bricken, Templeton, Batson, Chen, Jermyn, Conerly,
  Turner, Anil, Denison, Askell, Lasenby, Wu, Kravec, Schiefer, Maxwell,
  Joseph, Tamkin, Nguyen, McLean, Burke, Hume, Carter, Henighan, and
  Olah]{bricken_towards_2023}
Trenton Bricken, Adly Templeton, Joshua Batson, Brian Chen, Adam Jermyn, Tom
  Conerly, Nicholas~L. Turner, Cem Anil, Carson Denison, Amanda Askell, Robert
  Lasenby, Yifan Wu, Shauna Kravec, Nicholas Schiefer, Tim Maxwell, Nicholas
  Joseph, Alex Tamkin, Karina Nguyen, Brayden McLean, Josiah~E. Burke, Tristan
  Hume, Shan Carter, Tom Henighan, and Chris Olah.
\newblock Towards {Monosemanticity}: {Decomposing} {Language} {Models} {With}
  {Dictionary} {Learning}, 2023.
\newblock URL
  \url{https://transformer-circuits.pub/2023/monosemantic-features}.

\bibitem[Jacovi and Goldberg(2020)]{jacovi_towards_2020}
Alon Jacovi and Yoav Goldberg.
\newblock Towards {Faithfully} {Interpretable} {NLP} {Systems}: {How} {Should}
  {We} {Define} and {Evaluate} {Faithfulness}?
\newblock In Dan Jurafsky, Joyce Chai, Natalie Schluter, and Joel Tetreault,
  editors, \emph{Proceedings of the 58th {Annual} {Meeting} of the
  {Association} for {Computational} {Linguistics}}, pages 4198--4205, Online,
  July 2020. Association for Computational Linguistics.
\newblock \doi{10.18653/v1/2020.acl-main.386}.
\newblock URL \url{https://aclanthology.org/2020.acl-main.386/}.

\bibitem[Gurnee(2024)]{gurnee_sae_2024}
Wes Gurnee.
\newblock {SAE} reconstruction errors are (empirically) pathological — {AI}
  {Alignment} {Forum}, March 2024.
\newblock URL
  \url{https://www.alignmentforum.org/posts/rZPiuFxESMxCDHe4B/sae-reconstruction-errors-are-empirically-pathological}.

\bibitem[Engels et~al.(2024)Engels, Smith, and
  Tegmark]{engels_decomposing_2024}
Joshua Engels, Logan~Riggs Smith, and Max Tegmark.
\newblock Decomposing {The} {Dark} {Matter} of {Sparse} {Autoencoders}.
\newblock \emph{Transactions on Machine Learning Research}, December 2024.
\newblock ISSN 2835-8856.
\newblock URL \url{https://openreview.net/forum?id=sXq3Wb3vef}.

\bibitem[Chanin et~al.(2025)Chanin, Wilken-Smith, Dulka, Bhatnagar, Golechha,
  and Bloom]{chanin_is_2025}
David Chanin, James Wilken-Smith, Tomáš Dulka, Hardik Bhatnagar, Satvik
  Golechha, and Joseph~Isaac Bloom.
\newblock A is for {Absorption}: {Studying} {Feature} {Splitting} and
  {Absorption} in {Sparse} {Autoencoders}.
\newblock October 2025.
\newblock URL \url{https://openreview.net/forum?id=R73ybUciQF}.

\bibitem[Gao et~al.(2026)Gao, Meng, Zhou, and Pan]{gao_towards_2026}
Yutong Gao, Qinglin Meng, Yuan Zhou, and Liangming Pan.
\newblock Towards {Intrinsic} {Interpretability} of {Large} {Language}
  {Models}:{A} {Survey} of {Design} {Principles} and {Architectures}, April
  2026.
\newblock URL \url{http://arxiv.org/abs/2604.16042}.
\newblock arXiv:2604.16042 [cs].

\bibitem[Herbst et~al.(2026)Herbst, Lee, and Wermter]{herbst_expert_2026}
Jeremy Herbst, Jae~Hee Lee, and Stefan Wermter.
\newblock The {Expert} {Strikes} {Back}: {Interpreting} {Mixture}-of-{Experts}
  {Language} {Models} at {Expert} {Level}, April 2026.
\newblock URL \url{http://arxiv.org/abs/2604.02178}.
\newblock arXiv:2604.02178 [cs].

\bibitem[Elhage et~al.(2022{\natexlab{a}})Elhage, Hume, Olsson, Nanda,
  Henighan, Johnston, Showk, Joseph, DasSarma, Mann, Hernandez, Askell,
  Ndousse, Jones, Drain, Chen, Bai, Ganguli, Lovitt, Hatfield-Dodds, Kernion,
  Conerly, Kravec, Fort, Kadavath, Jacobson, Tran-Johnson, Kaplan, Clark,
  Brown, McCandlish, Amodei, and June~27]{elhage_softmax_2022}
Nelson Elhage, Tristan Hume, Catherine Olsson, Neel Nanda, Tom Henighan, Scott
  Johnston, Sheer~El Showk, Nicholas Joseph, Nova DasSarma, Ben Mann, Danny
  Hernandez, Amanda Askell, Kamal Ndousse, Andy Jones, Dawn Drain, Anna Chen,
  Yuntao Bai, Deep Ganguli, Liane Lovitt, Zac Hatfield-Dodds, Jackson Kernion,
  Tom Conerly, Shauna Kravec, Stanislav Fort, Saurav Kadavath, Josh Jacobson,
  Eli Tran-Johnson, Jared Kaplan, Jack Clark, Tom Brown, Sam McCandlish, Dario
  Amodei, and Christopher Olah‡ AFFILIATION Anthropic~PUBLISHED June~27.
\newblock Softmax {Linear} {Units}, 2022{\natexlab{a}}.
\newblock URL \url{https://transformer-circuits.pub/2022/solu/index.html}.

\bibitem[Sharkey(2023)]{sharkey_technical_2023}
Lee Sharkey.
\newblock A technical note on bilinear layers for interpretability, May 2023.
\newblock URL \url{http://arxiv.org/abs/2305.03452}.
\newblock arXiv:2305.03452 [cs].

\bibitem[Elhage et~al.(2021)Elhage, Nanda, Olsson, Henighan, Joseph, Mann,
  Askell, Bai, Chen, Conerly, DasSarma, Drain, Ganguli, Hatfield-Dodds,
  Hernandez, Jones, Kernion, Lovitt, Ndousse, Amodei, Brown, Clark, Kaplan,
  McCandlish, and Olah]{elhage_mathematical_2021}
Nelson Elhage, Neel Nanda, Catherine Olsson, Tom Henighan, Nicholas Joseph, Ben
  Mann, Amanda Askell, Yuntao Bai, Anna Chen, Tom Conerly, Nova DasSarma, Dawn
  Drain, Deep Ganguli, Zac Hatfield-Dodds, Danny Hernandez, Andy Jones, Jackson
  Kernion, Liane Lovitt, Kamal Ndousse, Dario Amodei, Tom Brown, Jack Clark,
  Jared Kaplan, Sam McCandlish, and Chris Olah.
\newblock A {Mathematical} {Framework} for {Transformer} {Circuits}, 2021.
\newblock URL \url{https://transformer-circuits.pub/2021/framework/index.html}.

\bibitem[Saxe et~al.(2014)Saxe, McClelland, and Ganguli]{saxe_exact_2014}
Andrew~M. Saxe, James~L. McClelland, and Surya Ganguli.
\newblock Exact solutions to the nonlinear dynamics of learning in deep linear
  neural networks, 2014.
\newblock URL \url{https://arxiv.org/abs/1312.6120}.
\newblock \_eprint: 1312.6120.

\bibitem[Saxe et~al.(2022)Saxe, Sodhani, and Lewallen]{saxe_neural_2022}
Andrew~M. Saxe, Shagun Sodhani, and Sam Lewallen.
\newblock The {Neural} {Race} {Reduction}: {Dynamics} of {Abstraction} in
  {Gated} {Networks}, July 2022.
\newblock URL \url{http://arxiv.org/abs/2207.10430}.
\newblock arXiv:2207.10430 [cs].

\bibitem[Elhage et~al.(2022{\natexlab{b}})Elhage, Hume, Olsson, Schiefer,
  Henighan, Kravec, Hatfield-Dodds, Lasenby, Drain, Chen, Grosse, McCandlish,
  Kaplan, Amodei, Wattenberg, and Olah]{elhage_toy_2022}
Nelson Elhage, Tristan Hume, Catherine Olsson, Nicholas Schiefer, Tom Henighan,
  Shauna Kravec, Zac Hatfield-Dodds, Robert Lasenby, Dawn Drain, Carol Chen,
  Roger Grosse, Sam McCandlish, Jared Kaplan, Dario Amodei, Martin Wattenberg,
  and Christopher Olah.
\newblock Toy {Models} of {Superposition}, 2022{\natexlab{b}}.
\newblock URL \url{https://transformer-circuits.pub/2022/toy_model/index.html}.

\bibitem[Bradbury et~al.(2018)Bradbury, Frostig, Hawkins, Johnson, Leary,
  Maclaurin, Necula, Paszke, VanderPlas, Wanderman-Milne, and
  Zhang]{bradbury_jax_2018}
James Bradbury, Roy Frostig, Peter Hawkins, Matthew~James Johnson, Chris Leary,
  Dougal Maclaurin, George Necula, Adam Paszke, Jake VanderPlas, Skye
  Wanderman-Milne, and Qiao Zhang.
\newblock {JAX}: composable transformations of {Python}+{NumPy} programs, 2018.
\newblock URL \url{http://github.com/google/jax}.

\bibitem[Heek et~al.(2023)Heek, Levskaya, Oliver, Ritter, Rondepierre, Steiner,
  and Zee]{heek_flax_2023}
Jonathan Heek, Anselm Levskaya, Avital Oliver, Marvin Ritter, Bertrand
  Rondepierre, Andreas Steiner, and Marc~van Zee.
\newblock Flax: {A} neural network library and ecosystem for {JAX}, 2023.
\newblock URL \url{http://github.com/google/flax}.

\bibitem[Babuschkin et~al.(2020)Babuschkin, Baumli, Bell, Bhupatiraju, Bruce,
  Buchlovsky, Budden, Cai, Clark, Danihelka, Dedieu, Fantacci, Godwin, Jones,
  Hemsley, Hennigan, Hessel, Hou, Kapturowski, Keck, Kemaev, King, Kunesch,
  Martens, Merzic, Mikulik, Norman, Papamakarios, Quan, Ring, Ruiz, Sanchez,
  Schneider, Sezener, Spencer, Srinivasan, Stokowiec, Wang, Zhou, and
  Viola]{babuschkin_deepmind_2020}
Igor Babuschkin, Kate Baumli, Alison Bell, Surya Bhupatiraju, Jake Bruce, Peter
  Buchlovsky, David Budden, Trevor Cai, Aidan Clark, Ivo Danihelka, Antoine
  Dedieu, Claudio Fantacci, Jonathan Godwin, Chris Jones, Ross Hemsley, Tom
  Hennigan, Matteo Hessel, Shaobo Hou, Steven Kapturowski, Thomas Keck, Iurii
  Kemaev, Michael King, Markus Kunesch, Lena Martens, Hamza Merzic, Vladimir
  Mikulik, Tamara Norman, George Papamakarios, John Quan, Roman Ring, Francisco
  Ruiz, Alvaro Sanchez, Rosalia Schneider, Eren Sezener, Stephen Spencer,
  Srivatsan Srinivasan, Wojciech Stokowiec, Luyu Wang, Guangyao Zhou, and Fabio
  Viola.
\newblock The {DeepMind} {JAX} {Ecosystem}, 2020.
\newblock URL \url{http://github.com/deepmind}.

\bibitem[Ritter et~al.(2023)Ritter, Indyk, Singh, Audibert, Seelam, Hanes, Lau,
  Olesiak, Kang, and Wu]{ritter_grain_2023}
Marvin Ritter, Ihor Indyk, Aayush Singh, Andrew Audibert, Anoosha Seelam,
  Camelia Hanes, Eric Lau, Jacek Olesiak, Jiyang Kang, and Xihui Wu.
\newblock Grain - {Feeding} {JAX} {Models}, 2023.
\newblock URL \url{http://github.com/google/grain}.

\bibitem[Lhoest et~al.(2021)Lhoest, Villanova~del Moral, Jernite, Thakur, von
  Platen, Patil, Chaumond, Drame, Plu, Tunstall, Davison, Šaško, Chhablani,
  Malik, Brandeis, Le~Scao, Sanh, Xu, Patry, McMillan-Major, Schmid, Gugger,
  Delangue, Matussière, Debut, Bekman, Cistac, Goehringer, Mustar, Lagunas,
  Rush, and Wolf]{lhoest_datasets_2021}
Quentin Lhoest, Albert Villanova~del Moral, Yacine Jernite, Abhishek Thakur,
  Patrick von Platen, Suraj Patil, Julien Chaumond, Mariama Drame, Julien Plu,
  Lewis Tunstall, Joe Davison, Mario Šaško, Gunjan Chhablani, Bhavitvya
  Malik, Simon Brandeis, Teven Le~Scao, Victor Sanh, Canwen Xu, Nicolas Patry,
  Angelina McMillan-Major, Philipp Schmid, Sylvain Gugger, Clément Delangue,
  Théo Matussière, Lysandre Debut, Stas Bekman, Pierric Cistac, Thibault
  Goehringer, Victor Mustar, François Lagunas, Alexander Rush, and Thomas
  Wolf.
\newblock Datasets: {A} {Community} {Library} for {Natural} {Language}
  {Processing}.
\newblock In Heike Adel and Shuming Shi, editors, \emph{Proceedings of the 2021
  {Conference} on {Empirical} {Methods} in {Natural} {Language} {Processing}:
  {System} {Demonstrations}}, pages 175--184, Online and Punta Cana, Dominican
  Republic, November 2021. Association for Computational Linguistics.
\newblock \doi{10.18653/v1/2021.emnlp-demo.21}.
\newblock URL \url{https://aclanthology.org/2021.emnlp-demo.21/}.

\bibitem[Wolf et~al.(2020)Wolf, Debut, Sanh, Chaumond, Delangue, Moi, Cistac,
  Rault, Louf, Funtowicz, Davison, Shleifer, Platen, Ma, Jernite, Plu, Xu,
  Scao, Gugger, Drame, Lhoest, and Rush]{wolf_transformers_2020}
Thomas Wolf, Lysandre Debut, Victor Sanh, Julien Chaumond, Clement Delangue,
  Anthony Moi, Pierric Cistac, Tim Rault, Rémi Louf, Morgan Funtowicz, Joe
  Davison, Sam Shleifer, Patrick~von Platen, Clara Ma, Yacine Jernite, Julien
  Plu, Canwen Xu, Teven~Le Scao, Sylvain Gugger, Mariama Drame, Quentin Lhoest,
  and Alexander~M. Rush.
\newblock Transformers: {State}-of-the-{Art} {Natural} {Language} {Processing}.
\newblock In \emph{Proceedings of the 2020 {Conference} on {Empirical}
  {Methods} in {Natural} {Language} {Processing}: {System} {Demonstrations}},
  pages 38--45, Online, October 2020. Association for Computational
  Linguistics.
\newblock URL \url{https://aclanthology.org/2020.emnlp-demos.6/}.

\bibitem[Pedregosa et~al.(2011)Pedregosa, Varoquaux, Gramfort, Michel, Thirion,
  Grisel, Blondel, Prettenhofer, Weiss, Dubourg, Vanderplas, Passos,
  Cournapeau, Brucher, Perrot, and Duchesnay]{pedregosa_scikit-learn_2011}
F.~Pedregosa, G.~Varoquaux, A.~Gramfort, V.~Michel, B.~Thirion, O.~Grisel,
  M.~Blondel, P.~Prettenhofer, R.~Weiss, V.~Dubourg, J.~Vanderplas, A.~Passos,
  D.~Cournapeau, M.~Brucher, M.~Perrot, and E.~Duchesnay.
\newblock Scikit-learn: {Machine} {Learning} in {Python}.
\newblock \emph{Journal of Machine Learning Research}, 12:\penalty0 2825--2830,
  2011.

\bibitem[Biewald(2020)]{biewald_experiment_2020}
Lukas Biewald.
\newblock Experiment {Tracking} with {Weights} and {Biases}, 2020.
\newblock URL \url{https://www.wandb.com/}.

\bibitem[Inc(2015)]{inc_collaborative_2015}
Plotly~Technologies Inc.
\newblock Collaborative data science, 2015.
\newblock URL \url{https://plot.ly}.
\newblock Place: Montreal, QC.

\bibitem[Marsh(2024)]{marsh_uv_2024}
Charlie Marsh.
\newblock uv: {An} extremely fast {Python} package and project manager, written
  in {Rust}., 2024.
\newblock URL \url{https://pypi.org/project/uv/}.

\end{thebibliography}
\bibliographystyle{unsrtnat}

\appendix
\newpage
\noindent\rule{\textwidth}{1pt}
\begin{center}
    {\LARGE \bf Appendix \par}
\end{center}
\startcontents[sections]
\printcontents[sections]{l}{1}{\setcounter{tocdepth}{3}}
\noindent\rule{\textwidth}{1pt}

\newpage
\section{Sparsely gated linear neuron layer}
\label{appsec:sgatlin-channels}

\subsection{Parallel channels}
The full \texttt{sgatlin} layer consists of $C$ parallel \textit{channels} that are summed,
\begin{align*}
  \label{eq:sgatlin_channels}
  \bm{g}_c &= \texttt{product-top-k}(\bm{W}_c^{\text{key}} \bm{W}^{\text{query}} \bm{z}),\\ %
  \texttt{sgatlin}(\bm{z}) &= \sum_{c=1}^C \sum_{i: g_{c,i} \neq 0} g_{c,i} \bm{w}_{c, i}^{\text{out}} \bm{w}_{c, i}^{\text{in}\top}\bm{z}
\end{align*}

where $\bm{W}^{\text{query}} \in \mathbb{R}^{d_{\text{key}} \times d_{\text{model}}}$ is shared across layers, whereas $\bm{W}^{\text{key}}_c \in \mathbb{R}^{2 \sqrt{d_{\text{ffw}}} \times d_{\text{key}} }$, $\bm{g}_{c}\in \mathbb{R}^{d_{\text{ffw}}}$, and $\bm{w}_{c, i}^{\text{in}}, \bm{w}_{c, i}^{\text{out}} \in \mathbb{R}^{d_{\text{model}}}$ are all channel-specific.
$\texttt{product-top-k}(\bm{x}): \mathbb{R}^{2 \sqrt{d_{\text{ffw}}}} \to \mathbb{R}^{d_{\text{ffw}}}$ is the product top-$k$ operator introduced by \citet{lample_large_2019} that selects the $k$ largest entries of the product set of two sets of scalars, here obtained by slicing the input vector $\bm{x}$ into two parts of equal size, and returns a sparse vector of gating scores with $k$ non-zero entries.
Pseudocode for the layer is shown in Listing~\ref{alg:sgatlin}.

\begin{listing}[htpb]
\begin{lstlisting}
class SparselyGatedLinear(nn.Module):
  d_model: int
  d_ffw: int
  d_key: int
  knn: int
  n_channels: int

  @nn.compact
  def __call__(self, x):

    # Gating network
    n_keys = int(math.sqrt(self.d_ffw))
    queries = nn.Dense(features=self.d_key, use_bias=False)(x)
    similar = nn.DenseGeneral(
      features=(2, self.n_channels, n_keys),
      use_bias=False,
    )(queries)
    scores, indices = product_key_top_k(similar, k=self.knn)

    # Expert computation
    experts = self.param(
      name='experts',
      init_fn=lecun_normal,
      shape=(2, self.n_channels, self.d_ffw, self.d_model),
    )

    w_in, w_out = experts[indices]
    z = jnp.einsum('btd,btckd->btck', x, w_in)
    z = z * scores
    z = jnp.einsum('btck,btckd->btd', z, w_out)

    return z
\end{lstlisting}
  \caption{Pseudocode for the \texttt{sgatlin} layer using JAX \citep{bradbury_jax_2018} and Flax-like \citep{heek_flax_2023} syntax.}
  \label{alg:sgatlin}
\end{listing}

\subsection{Time and memory complexity}
We compare the time and memory complexities of different transformer feedforward layers in Table~\ref{tab:memory-time-complexity-feedforward}.
In typical dense feedforward layers, the time complexity is dominated by the product of the residual stream dimension and the feedforward dimension.
Scaling the number of parameters by increasing the feedforward dimension proportionally increases computational cost.
MoEs decouple the total number of feedforward neurons (i.e., the product of the number of experts and the size of each expert) from the number of active feedforward neurons by activating only a subset of experts.
This allows scaling the number of feedforward parameters without proportionally increasing computational cost.
However, typical MoE parameterizations are usually comparably coarse-grained, and the size of each expert is large enough for the time complexity to be dominated by the product of active experts and their size.
Both PEER \citep{he_mixture_2024} and \texttt{sgatlin} shrink the experts to consist of a single neuron and only activate a small, fixed subset of neurons per token.
In combination with their respective efficient router/gating network, this allows scaling the number of parameters while keeping the time complexity comparably small.
Note that the difference in time and memory complexity between PEER and \texttt{sgatlin} is due to the gating network query projection being shared across channels in the latter but not in the former.

\begin{table}[t]
\centering
\caption{\textbf{Time and memory complexities of feedforward layers}. We compare the time and memory complexities of different transformer feedforward layers considered in our experiments. $d_{\text{model}}$ denotes the residual stream dimension, $d_{\text{ffw}}$ is the number of feedforward neurons, $n_{\text{experts}}$ the total number of experts, $n_{\text{active}}$ the number of active experts routed per token, $n_{\text{channel}}$ the number of channels, and $d_{\text{key}}$ the bottleneck dimension of the gating network.}
\label{tab:memory-time-complexity-feedforward}
\vspace{0.5em}
\renewcommand{\arraystretch}{1.3}
\resizebox{\textwidth}{!}{
\begin{tabular}{lll}
\toprule
\textbf{Feedforward variant} & \textbf{Time complexity} ($\mathcal{O}$) & \textbf{Memory complexity} ($\mathcal{O}$) \\ \midrule
MLP & $d_{\text{model}} \, d_{\text{ffw}}$ & $d_{\text{model}} \, d_{\text{ffw}}$ \\
SwiGLU & $d_{\text{model}} \, d_{\text{ffw}}$ & $d_{\text{model}} \, d_{\text{ffw}}$ \\
MoE & $d_{\text{model}} \, n_{\text{experts}} + n_{\text{active}} \, d_{\text{model}} \, d_{\text{ffw}}$ & $n_{\text{experts}} \, d_{\text{model}} \, d_{\text{ffw}}$ \\
PEER & $n_{\text{channel}} \, d_{\text{key}} (d_{\text{model}} + \sqrt{d_{\text{ffw}}} ) + n_{\text{channel}} \, d_{\text{model}} \, n_{\text{active}} $ & $n_{\text{channel}} \, d_{\text{key}} (d_{\text{model}} + \sqrt{d_{\text{ffw}}}) + d_{\text{ffw}} \, d_{\text{model}}$ \\ \midrule
\texttt{sgatlin} & $d_{\text{key}} ( d_{\text{model}} + n_{\text{channel}} \sqrt{d_{\text{ffw}}})  + n_{\text{channel}} \, d_{\text{model}} \, n_{\text{active}} $ & $d_{\text{key}} ( d_{\text{model}} + n_{\text{channel}} \sqrt{d_{\text{ffw}}}) + n_{\text{channel}} \, d_{\text{ffw}} \, d_{\text{model}}$ \\ \bottomrule
\end{tabular}
}
\end{table}

\section{Experimental details}
\label{appsec:experimental-details}

\subsection{Training details}
\label{appsec:training-details}

In the following, we provide further training details for both the language modelling experiments on the SlimPajama 627B dataset \citep{soboleva_slimpajama_2023} conducted in Section~\ref{sec:language-modelling} and the model interpretability experiments based on the TinyStories dataset \citep{eldan_tinystories_2023} conducted in Section~\ref{sec:model-interpretability}.

\subsubsection{SlimPajama}

\paragraph{Tokenizer}
We tokenize text input using the pretrained GPT-2 byte-pair encoding tokenizer with  a vocabulary size of $\num{50257}$.

\paragraph{Model}
We use a causal, decoder-only transformer with an RMSNorm applied prior to the attention and feedforward block and rotary positional encoding.
We follow the common procedure of fixing the attention head dimension to 64 across all model scales and determining the number of attention heads according to the residual stream dimension of the respective experiment.
We use the MoE layer from GPT-OSS \citep{openai_gpt-oss-120b_2025} with 16 experts and top-$k=2$ routing.
Both PEER and \texttt{sgatlin} use 16 channels, top-$k=8$ gating and a gating bottleneck dimension of $128$.
All models are trained in \texttt{bfloat16} precision with \texttt{float32} upcasting on the attention's softmax operation and the final output embedding.
In addition, we use activation checkpointing (\texttt{flax.linen.remat}) to further reduce the memory requirements during training.

\paragraph{Optimizer}
We use a global batch size of $128$ and a sequence length of $2048$ tokens.
We optimize the models using AdamW with a peak learning rate of $\num{1e-3}$ and apply a weight decay of $0.1$, excluding scalar parameters.
The learning rate follows a warmup stable square root decay schedule \citep{hu_minicpm_2024,hagele_scaling_2024} with $1000$ warmup steps and a decay fraction of 0.2.
Gradients are clipped to a maximum global norm of 1.0.
Learning rate, weight decay, and warmup steps were selected after running a grid search over learning rate values of $\{\num{1e-4}, \num{3e-4}, \num{1e-3}\}$, weight decay values of $\{0.01, 0.03, 0.1\}$, and warmup steps of $\{100, 1000\}$, training a dense SwiGLU transformer with 12 layers on a compute budget of $\num{1e18}$~FLOPs for each configuration.

\subsubsection{TinyStories}
\label{appsec:hyperparameters-tinystories}

For our model interpretability experiments with the TinyStories dataset, we train \texttt{sgatlin}-based transformers with 4 layers, a residual stream dimension of 256 and 4 attention heads of dimension 64 with a batch size of $256$ sequences and a sequence length of $1024$ tokens.
We train a custom byte-pair encoding tokenizer on the TinyStories training data with a vocabulary size of $8192$.
Otherwise, we reuse the same optimizer and scheduler settings as for SlimPajama, only reducing the number of warmup steps to $100$ and increasing the gradient clipping norm to 2.0.

\subsection{Scaling ladder}
\label{appsec:scaling-ladder}

We use the scaling ladder shown in Listing~\ref{alg:scaling-ladder} to systematically increase the depth and width of our models.
In our experiments, we vary the \texttt{scale} integer from 2 to 8.
\begin{listing}[h]
\begin{lstlisting}
def scaling_ladder(scale: int, feedforward_type: str):
  d_model = 128 * scale
  n_layers = 2 * scale

  match feedforward_type:
    case 'swiglu' | 'mlp':
      d_ffw = int(8 / 3 * d_model / 256) * 256
    case 'moe':
      d_ffw = d_model  # following GPT-OSS
    case 'peer':
      d_ffw = (32 + 24 * scale) ** 2
    case 'sgatlin':
      d_ffw = (16 + 12 * scale) ** 2
    case _:
      raise ValueError(f'{feedforward_type} unknown')

  return (d_model, n_layers, d_ffw)
\end{lstlisting}
  \caption{Python code for scaling transformer model size with different feedforward layer types.}
  \label{alg:scaling-ladder}
\end{listing}

\section{Additional results}
\subsection{Neuron usage and coactivation}

\begin{figure}
  \includegraphics[width=0.52\textwidth]{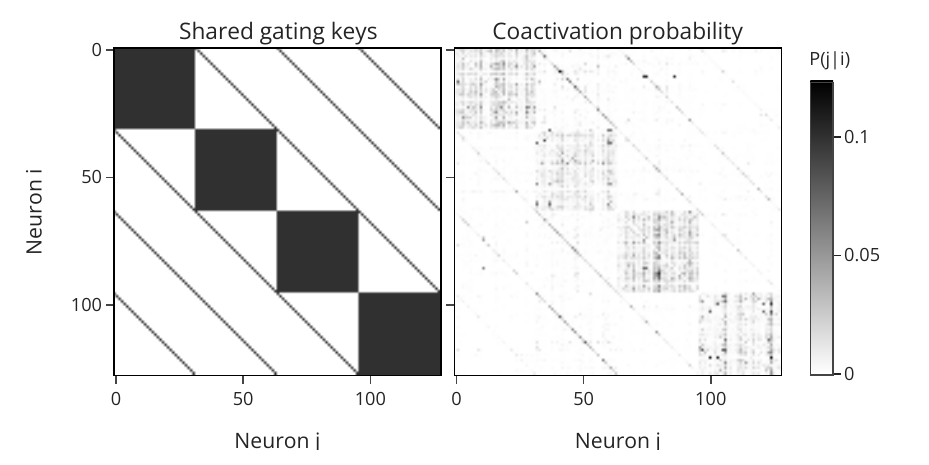}
  \includegraphics[width=0.20\textwidth]{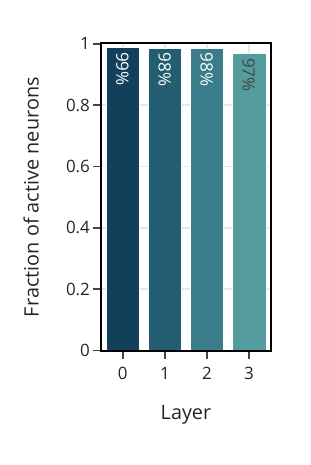}
  \includegraphics[width=0.27\textwidth]{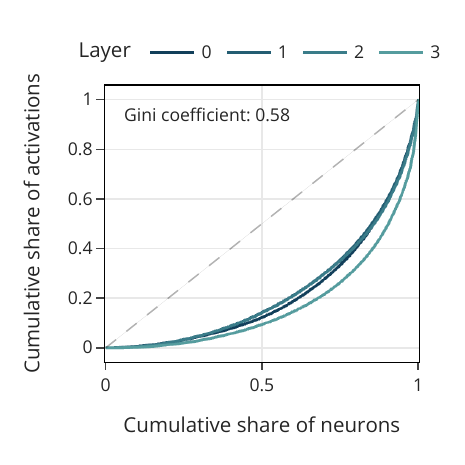}
  \caption{\textbf{Neuron usage and coactivation.} \textbf{\textit{Left}} As a result of the product top-$k$ mechanism, certain pairs of neurons share gating keys. This affects the empirically observed coactivation probability of neurons, where neurons preferably coactivate with neurons that they share a gating key with. Coactivation probability is computed over 128 random sequences of the TinyStories validation set and shown for the first 128 neurons of the penultimate transformer layer. \textbf{\textit{Center}} Almost all neurons activate at least once over 128 random sequences of the TinyStories validation set. \textbf{\textit{Right}} Neuron activation frequency is heterogeneous, with an average Gini coefficient of 0.58 across 128 random sequences of the TinyStories validation set. We control for the skewed input token frequency of TinyStories sequences considering each unique input token only once.}
  \label{fig:neuron-usage-coactivation}
\end{figure}

In Figure~\ref{fig:neuron-usage-coactivation}, we show neuron usage and coactivation in the \texttt{sgatlin}-based transformer trained on TinyStories used for our model interpretability experiments.

\paragraph{The product top-$k$ mechanism creates neuron subpopulations.}
The product top-$k$ mechanism used in the gating network of \texttt{sgatlin} computes an input-dependent query and calculates the dot-product with a neuron-specific key.
It efficiently computes the dot-product for a large number of possible neurons by defining the neuron keys as the product-set of multiple subkeys.
As a result, each neuron shares its two subkeys with other neurons.
This is illustrated in the left panel of Figure~\ref{fig:neuron-usage-coactivation}.
In practice, this key sharing affects the empirically observed coactivation probability of neurons, effectively creating subpopulations of neurons that preferably coactivate with neurons they share a gating key with.

\paragraph{Available neurons are effectively used.}
We verify that the gating network effectively uses the available feedforward capacity and activates almost all the available feedforward neurons at least once across a batch of $128$ randomly chosen sequences from the TinyStories validation set, as shown in the centre panel of Figure~\ref{fig:neuron-usage-coactivation}.
The right panel further shows that neuron activation frequency is heterogeneous, with an average Gini coefficient of 0.58 across these sequences.

\subsection{Model interpretability}

\begin{figure}
\centering
  \includegraphics[width=\textwidth]{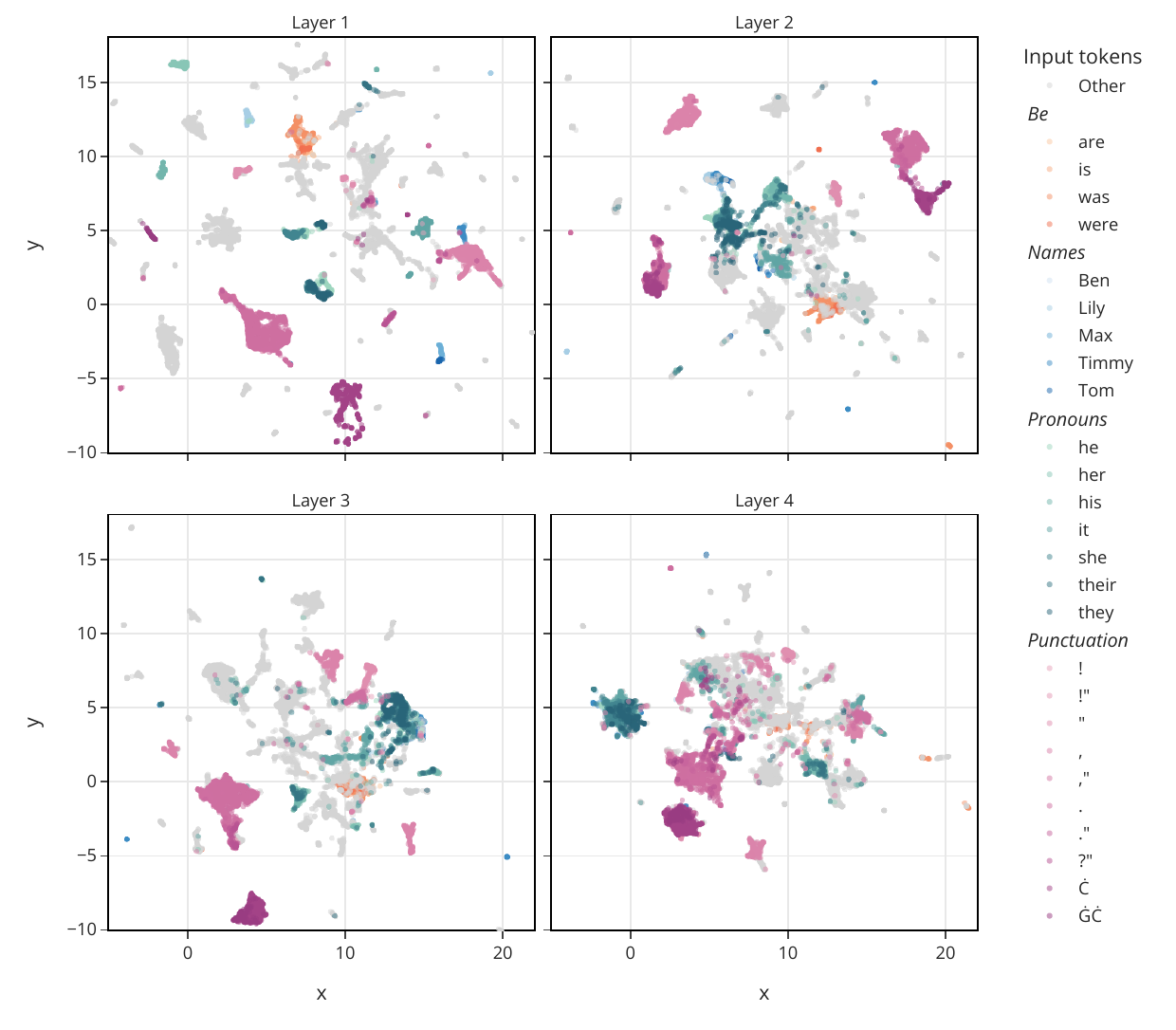}
  \vspace{-1em}
  \caption{\textbf{\texttt{sgatlin} gating weights form semantic clusters (extended version of Figure~\ref{fig:gating-embedding}).} We collect the gating weights of \texttt{sgatlin} of a four layer language model trained on TinyStories for the 100 most frequent tokens and embed them into 2D space using UMAP. The resulting embedding forms clusters that are partially explained by the semantics of the corresponding input tokens. For instance, names used in the stories cluster together and pronouns form clusters in their vicinity.}
  \label{fig:gating-embedding-all-layers}
\end{figure}

Figure~\ref{fig:gating-embedding-all-layers} shows an extended version of Figure~\ref{fig:gating-embedding} in the main text, with gating embeddings for all four layers.
Table~\ref{tab:gating-neighbors-extended} shows an extended version of Figure~\ref{fig:gating-neighbors} in the main text, contextualizing the feedforward circuit neighbours with excerpts from the input sequences that activated them as well as the top 5 output predictions at that respective token position.

\begin{landscape}
\begin{longtable}{lllll}
\caption{\textbf{Feedforward circuit neighbours (extended).} Following the procedure shown in Figure~\ref{fig:circuits-conceptual} of collecting reference gating weights across input sequences, we can analyse the feedforward circuits for a new input sequence by comparing them to their nearest neighbours among the reference set. In this example, we analyse the feedforward circuits at the penultimate layer for the new input sentence excerpt "\texttt{... big dog named Max. Max had a red collar that he wore every day}". The \textbf{bold} token in the 'Neighbour input excerpt' column denotes the token at which the neighbour circuit was activated. The 'Neighbour top predictions' column lists the 5 top output predictions at the output layer.}
\label{tab:gating-neighbors-extended} \\
\toprule
Pos & Token & Neighbour distance & Neighbour input excerpt & Neighbour top predictions \\
\midrule
\endfirsthead
\multicolumn{5}{c}{{\tablename\ \thetable{} -- Continued from previous page}} \\
\toprule
Pos & Token & Neighbour distance & Neighbour input excerpt & Neighbour top predictions \\
\midrule
\endhead
\midrule
\multicolumn{5}{r}{Continued on next page...} \\
\midrule
\endfoot
\bottomrule
\endlastfoot
9 &  big & 5.96e-08 &  time, there was a\textbf{ big} farm. On the farm & , | elephant | bear | dog | red \\
 &  & 5.96e-08 &  time, there was a\textbf{ big} and happy family who loved & , | elephant | bear | dog | red \\
 &  & 5.96e-08 &  time, there was a\textbf{ big} teddy bear. He loved & , | elephant | bear | dog | red \\
 &  & 5.96e-08 &  time, there was a\textbf{ big}, strong robot made of & , | elephant | bear | dog | red \\
 &  & 1.51e-01 &  time, there was an\textbf{ old} man. He liked to & man | lady | woman | house | tree \\
 &  & 1.51e-01 &  time, there was an\textbf{ old} lady. She liked to & man | lady | woman | house | tree \\
 &  & 2.06e-01 &  a queen who lays\textbf{ eggs} and makes more ants." & and | . | for | in | , \\
 &  & 2.20e-01 &  backyard where there was a\textbf{ long} fence. One day, & rope | vine | fence | hose | , \\
 &  & 2.23e-01 & . Amy had a big\textbf{ yard} with a purple swing. & with | where | . | in | to \\
 &  & 2.38e-01 &  Inside the boat was a\textbf{ little} boy called Ben. Ben & girl | boy | fish | boat | bird \\
\midrule
10 &  dog & 4.13e-02 & , there was a little\textbf{ dog} named Spot. Spot loved & named | . | called | who | gie \\
 &  & 4.92e-02 &  time, there was a\textbf{ dog} named Max. Max was & named | . | called | who | with \\
 &  & 5.57e-02 &  upon a time, a\textbf{ dog} named Max went for a & named | and | was | called | went \\
 &  & 6.44e-02 & , there was a sweet\textbf{ dog} named Max. Max loved & named | . | called | who | and \\
 &  & 1.11e-01 & , there was a little\textbf{ bunny} named Benny. Benny had & named | . | who | called | and \\
 &  & 1.32e-01 & , there was a little\textbf{ kitten} named Mittens. Mittens was & named | . | who | called | , \\
 &  & 1.38e-01 & , there was a graceful\textbf{ cat} named Kitty. She loved & named | . | who | called | and \\
 &  & 1.62e-01 &  Once there was a clever\textbf{ kitten}. Every day it would & . | named | who | called | , \\
 &  & 1.67e-01 & , there was a small\textbf{ mouse} named Timmy. Timmy was & named | . | who | called | that \\
 &  & 1.71e-01 &  was a big, strong\textbf{ robot} made of steel. He & . | named | who | called | that \\
\midrule
11 &  named & 3.18e-01 & , there was a dog\textbf{ named} Max. Max was a & Max | Spot | Buddy | Bob | Sam \\
 &  & 3.32e-01 &  there was a little dog\textbf{ named} Spot. Spot loved to & Spot | Max | Buddy | Lucky | Tim \\
 &  & 3.37e-01 &  there was a sweet dog\textbf{ named} Max. Max loved to & Max | Spot | Buddy | Daisy | Fluffy \\
 &  & 3.54e-01 &  a time, a dog\textbf{ named} Max went for a walk & Max | Spot | Buddy | Sam | Bow \\
 &  & 3.97e-01 &  there was a small mouse\textbf{ named} Timmy. Timmy was always & Timmy | Max | Tim | Mimi | Mickey \\
 &  & 4.03e-01 &  Mia met a talking cat\textbf{ named} Tom. Tom was a & Tom | Whiskers | Fluffy | Luna | Sam \\
 &  & 4.18e-01 &  named Lily and a boy\textbf{ named} Tom were playing in their & Tom | Tim | Max | Ben | Sam \\
 &  & 4.48e-01 &  She had a pet cat\textbf{ named} Olive. Lily loved to & Mittens | Fluffy | Whiskers | Tom | Max \\
 &  & 5.01e-01 &  a puppy named Max who\textbf{ was} her best friend. One & very | her | always | a | so \\
 &  & 5.21e-01 &  this is my plant.\textbf{ You} can't touch it." & can | are | cannot | have | will \\
\midrule
12 &  Max & 1.13e-02 &  there was a dog named\textbf{ Max}. Max was a wild & . | who | and | , | the \\
 &  & 2.83e-02 &  was a little dog named\textbf{ Spot}. Spot loved to play & . | who | and | , | ty \\
 &  & 2.91e-02 &  was a sweet dog named\textbf{ Max}. Max loved to jog & . | who | and | , | ! \\
 &  & 4.29e-02 &  there were two birds called\textbf{ Jack} and Jill. Every night & and | . | , | who | 's \\
 &  & 4.66e-02 &  time, a dog named\textbf{ Max} went for a walk in & was | lived | went | and | wanted \\
 &  & 4.82e-02 &  was a little girl named\textbf{ Amy}. Amy had a big & . | who | and | , | a \\
 &  & 5.92e-02 &  was a little kitten named\textbf{ Mittens}. Mittens was very popular & . | who | , | and | that \\
 &  & 6.47e-02 &  there was a girl named\textbf{ Mia}. Mia loved her jewelry & . | who | and | a | , \\
 &  & 7.08e-02 &  was a graceful cat named\textbf{ Kitty}. She loved to play & . | who | , | and | ty \\
 &  & 7.12e-02 &  to a little boy named\textbf{ Jack}. He loved the fat & . | who | and | , | 's \\
\midrule
13 & . & 6.31e-03 &  a little boy called Ben\textbf{.} Ben was just 3 years & Ben | He | Ċ | One | Ċ \\
 &  & 7.51e-03 &  a little squirrel named Sammy\textbf{.} Sammy loved to play and & Sammy | He | One | Timmy | Nutty \\
 &  & 5.01e-02 &  was a dog named Max\textbf{.} Max was a wild dog & Max | He | One | His | It \\
 &  & 5.99e-02 &  a small mouse named Timmy\textbf{.} Timmy was always scared of & Timmy | He | One | Every | She \\
 &  & 6.07e-02 &  a little bunny named Benny\textbf{.} Benny had soft white fur & Benny | He | One | Every | Today \\
 &  & 6.25e-02 &  a little fish named Bubbles\textbf{.} Bubbles lived in a big & Bubbles | He | One | She | Nemo \\
 &  & 6.51e-02 &  deer named Bambi\textbf{.} Bambi wanted to & B | She | He | One | Every \\
 &  & 6.71e-02 &  a little boy named Jack\textbf{.} He loved the fat car & He | Jack | One | Every | Ċ \\
 &  & 7.90e-02 &  a little kitten named Mittens\textbf{.} Mittens was very popular with & Mittens | She | One | He | Mittens \\
 &  & 9.76e-02 &  was just 3 years old\textbf{.} He was sailing all by & Ċ | He | Ċ | One | Every \\
\midrule
14 &  Max & 1.02e-02 &  a girl named Mia.\textbf{ Mia} loved her jewelry. She & was | loved | had | liked | wanted \\
 &  & 1.42e-02 &  a dog named Max.\textbf{ Max} was a wild dog who & was | loved | had | liked | lived \\
 &  & 2.15e-02 &  little dog named Spot.\textbf{ Spot} loved to play fetch with & loved | was | had | liked | lived \\
 &  & 3.63e-02 &  little girl named Amy.\textbf{ Amy} had a big yard with & loved | had | was | liked | lived \\
 &  & 5.22e-02 &  sweet dog named Max.\textbf{ Max} loved to jog in the & loved | liked | had | was | lived \\
 &  & 7.19e-02 &  cakes. One day,\textbf{ she} went for a walk and & decided | wanted | went | baked | was \\
 &  & 9.46e-02 &  else. One day,\textbf{ Tim} saw a big candy store & went | 's | saw | found | was \\
 &  & 9.68e-02 &  friends. One day,\textbf{ he} heard a little girl crying & saw | was | found | went | heard \\
 &  & 1.03e-01 &  princess. One day,\textbf{ she} found a shiny belt in & found | decided | put | was | saw \\
 &  & 1.14e-01 &  good boy named Timmy.\textbf{ Timmy} always listened to his mommy & loved | liked | was | had | lived \\
\midrule
15 &  had & 1.61e-01 &  bunny named Benny. Benny\textbf{ had} soft white fur that was & a | soft | long | big | an \\
 &  & 1.97e-01 &  girl named Amy. Amy\textbf{ had} a big yard with a & a | an | long | many | two \\
 &  & 2.11e-01 &  boy named Tim. Tim\textbf{ had} a big, thin sack & a | an | long | very | many \\
 &  & 2.18e-01 &  girl named Lily. She\textbf{ had} a big wardrobe with smooth & a | long | an | to | very \\
 &  & 2.38e-01 &  girl named Lily. She\textbf{ had} a pet cat named Olive & a | long | an | many | lots \\
 &  & 2.69e-01 &  girl named Lily. Lily\textbf{ had} a tough day at school & a | an | long | to | very \\
 &  & 2.79e-01 &  there was a king who\textbf{ had} a lot of money. & a | to | an | very | lots \\
 &  & 2.90e-01 &  girl named Lily. She\textbf{ had} a puppy named Max who & a | an | long | to | very \\
 &  & 2.99e-01 &  dress-up. She\textbf{ had} a unique dress that her & a | many | lots | an | all \\
 &  & 3.11e-01 &  loved her jewelry. She\textbf{ had} a big box full of & a | many | lots | pretty | all \\
\midrule
16 &  a & 1.25e-01 &  named Tim. Tim had\textbf{ a} big, thin sack that & toy | big | red | new | friend \\
 &  & 1.40e-01 &  named Amy. Amy had\textbf{ a} big yard with a purple & big | toy | pretty | pet | favorite \\
 &  & 1.50e-01 &  out if the cat has\textbf{ a} home already. Maybe it & cut | bandage | home | thorn | boo \\
 &  & 1.63e-01 &  to read. He had\textbf{ a} big book with lots of & big | favorite | book | toy | red \\
 &  & 1.82e-01 &  named Lily. She had\textbf{ a} big wardrobe with smooth doors & big | pretty | toy | special | beautiful \\
 &  & 2.01e-01 &  a nice man who had\textbf{ a} red shirt. He liked & big | very | wife | special | lot \\
 &  & 2.03e-01 &  named Lily. She had\textbf{ a} puppy named Max who was & big | favorite | toy | pretty | pet \\
 &  & 2.26e-01 &  named Timmy. Timmy had\textbf{ a} red bike with a big & toy | big | favorite | pet | red \\
 &  & 2.26e-01 &  named Timmy. Timmy had\textbf{ a} big pocket on his shirt & toy | big | favorite | pet | red \\
 &  & 2.37e-01 &  named Lily. Lily had\textbf{ a} tough day at school because & big | toy | favorite | pet | pretty \\
\midrule
17 &  red & 2.19e-02 &  nice man who had a\textbf{ red} shirt. He liked to & car | wagon | hat | ball | coat \\
 &  & 1.28e-01 &  Timmy. Timmy had a\textbf{ red} bike with a big wheel & wagon | jacket | shirt | ball | vest \\
 &  & 1.28e-01 &  box. It has a\textbf{ red} light and a number. & button | mark | cross | label | bow \\
 &  & 1.37e-01 &  The different boy had a\textbf{ red} ball, and Timmy wanted & hat | shirt | ball | coat | jacket \\
 &  & 1.47e-01 &  a big yard with a\textbf{ purple} swing. She loved to & swing | fence | roof | ball | tree \\
 &  & 1.49e-01 &  Lily. She had a\textbf{ big} wardrobe with smooth doors. & , | box | sister | wardrobe | red \\
 &  & 1.60e-01 &  Amy. Amy had a\textbf{ big} yard with a purple swing & , | box | toy | red | dream \\
 &  & 1.65e-01 &  Timmy. Timmy had a\textbf{ big} pocket on his shirt. & , | sister | red | toy | box \\
 &  & 1.89e-01 &  Tim. Tim had a\textbf{ big}, thin sack that he & , | toy | box | red | sister \\
 &  & 2.25e-01 &  very rich and had a\textbf{ big} castle. One day, & house | castle | palace | garden | crown \\
\midrule
18 &  collar & 1.21e-01 & . She had a unique\textbf{ dress} that her mom made just & that | with | and | , | made \\
 &  & 1.28e-01 &  man who had a red\textbf{ shirt}. He liked to wear & . | and | that | on | with \\
 &  & 1.42e-01 &  a big wardrobe with smooth\textbf{ doors}. She loved to play & . | and | that | inside | where \\
 &  & 1.62e-01 &  red bike with a big\textbf{ wheel} in the front. He & . | that | on | and | in \\
 &  & 1.77e-01 &  brave knight with a colorful\textbf{ shield} and a sharp spear. & . | and | to | on | , \\
 &  & 1.82e-01 &  big yard with a purple\textbf{ swing}. She loved to play & . | that | in | and | on \\
 &  & 1.88e-01 &  big and had a shiny\textbf{ label} that said "brilliant & on | that | . | with | , \\
 &  & 1.98e-01 &  small toy car with a\textbf{ zipper} on it. He can & . | on | and | around | that \\
 &  & 2.04e-01 & . Benny had soft white\textbf{ fur} that was perfect for snugg & that | and | , | with | coat \\
 &  & 2.05e-01 &  different boy had a red\textbf{ ball}, and Timmy wanted to & and | that | , | . | in \\
\midrule
19 &  that & 2.94e-01 &  He had a white basketball\textbf{ that} he played with every day & he | was | his | made | bounced \\
 &  & 3.04e-01 &  She had a unique dress\textbf{ that} her mom made just for & she | her | was | made | sparkled \\
 &  & 3.18e-01 &  a big, thin sack\textbf{ that} he carried everywhere he went & he | his | was | could | kept \\
 &  & 3.20e-01 &  They had a small plant\textbf{ that} they watered every day. & they | grew | their | was | looked \\
 &  & 3.40e-01 &  they found a toy car\textbf{ that} Max had forgotten to put & they | was | Lily | looked | had \\
 &  & 3.77e-01 &  Tom found a big shell\textbf{ that} was shiny and had many & was | looked | had | he | sparkled \\
 &  & 4.14e-01 &  Benny had soft white fur\textbf{ that} was perfect for snuggling & he | was | felt | covered | made \\
 &  & 4.82e-01 &  It was the same shell\textbf{ that} she had found yesterday and & Tom | her | she | Mom | mom \\
 &  & 4.85e-01 & . She had colorful wings\textbf{ that} looked like a pastel drawing & were | sparkled | she | shone | helped \\
 &  & 5.12e-01 &  a big, impressive dog\textbf{ that} could do many tricks. & could | lived | came | made | barked \\
\midrule
20 &  he & 1.65e-01 &  big, thin sack that\textbf{ he} carried everywhere he went. & loved | liked | used | always | wanted \\
 &  & 1.80e-01 &  was the same shell that\textbf{ she} had found yesterday and left & had | found | saw | used | was \\
 &  & 2.16e-01 &  had a white basketball that\textbf{ he} played with every day. & played | would | always | liked | loved \\
 &  & 3.03e-01 &  had a puppy named Max\textbf{ who} was her best friend. & was | loved | she | always | liked \\
 &  & 4.32e-01 &  a different boy there who\textbf{ he} had never seen before. & didn | wanted | had | was | did \\
 &  & 4.83e-01 & , there was a shrimp\textbf{ named} Sammy. Sammy lived in & Sam | Tom | Tim | Jack | Lily \\
 &  & 5.05e-01 &  saw a different boy there\textbf{ who} he had never seen before & looked | was | didn | wanted | had \\
 &  & 5.07e-01 &  "It's a tool\textbf{ we} use to measure things. & use | can | need | have | measure \\
 &  & 5.11e-01 &  found a toy car that\textbf{ Max} had forgotten to put away & had | could | loved | wanted | was \\
 &  & 5.11e-01 &  daddy. They packed everything\textbf{ they} needed in a big, & needed | need | packed | would | had \\
\midrule
21 &  wore & 4.29e-01 &  there was a bigger glass\textbf{ of} milk with a straw set & milk | jam | water | sugar | sweet \\
 &  & 4.65e-01 &  big, impressive dog that\textbf{ could} do many tricks.Ċ & run | dance | do | talk | jump \\
 &  & 5.19e-01 & , impressive dog that could\textbf{ do} many tricks.ĊĊ & many | tricks | big | amazing | magic \\
 &  & 5.47e-01 &  see a metal thing that\textbf{ looks} like a stick. It & like | fun | very | old | interesting \\
 &  & 5.54e-01 &  had sent her the package\textbf{ of} pink stuff. Clara smiled & stuff | her | things | toys | fun \\
 &  & 5.60e-01 &  plant was hurt. Some\textbf{ of} the leaves and flowers fell & the | it | them | its | their \\
 &  & 5.81e-01 &  loved her jewelry. She\textbf{ had} a big box full of & a | many | lots | pretty | all \\
 &  & 5.85e-01 &  find any comfortable clothes to\textbf{ wear}. He tried on many & . | to | on | and | in \\
 &  & 6.11e-01 &  and strong. He protect\textbf{s} the castle from bad guys & us | me | the | our | my \\
 &  & 6.13e-01 & , they gave the bottle\textbf{ of} wine to the nice lady & grape | wine | juice | grapes | the \\
\midrule
22 &  every & 1.75e-01 &  basketball that he played with\textbf{ every} day. One day, & day | time | night | morning | week \\
 &  & 1.91e-01 &  small plant that they watered\textbf{ every} day. The plant had & day | night | morning | week | year \\
 &  & 3.25e-01 &  Lily loved to wear it\textbf{ every} day. ĊĊOne & day | night | time | morning | single \\
 &  & 3.32e-01 &  friends. They played together\textbf{ every} day.ĊĊOne & day | time | morning | week | Wednesday \\
 &  & 3.35e-01 &  so much to explore.\textbf{ Every} corner was filled with surprises & time | room | now | day | corner \\
 &  & 3.42e-01 &  eggs. She laid eggs\textbf{ every} day in her cozy nest & day | morning | night | year | week \\
 &  & 3.54e-01 &  their new puppy played together\textbf{ every} day and were the happiest & day | time | chance | week | weekend \\
 &  & 3.90e-01 &  and determination you can overcome\textbf{ any} challenge.<|endoftext|> & problem | challenge | ob | situation | difficult \\
 &  & 3.91e-01 &  friends and he waved his\textbf{ little} arms to say thank you & shrimp | legs | shell | friends | crab \\
 &  & 3.96e-01 &  in tip top shape.\textbf{ Every} day, she was happy & day | time | fish | morning | night \\
\midrule
23 &  day & 2.26e-01 & And I will protect the\textbf{ castle} with you and your dragon & from | ." | . | , | and \\
 &  & 2.67e-01 &  was friendly and licked her\textbf{ face}. Lily petted him and & . | to | , | with | ! \\
 &  & 2.93e-01 &  She laughed and ate more\textbf{ pie}.ĊĊFrom then & . | , | until | and | with \\
 &  & 3.18e-01 &  and enjoyed their tea and\textbf{ cookies}.ĊĊSuddenly, & . | together | while | , | with \\
 &  & 3.28e-01 &  not to play in the\textbf{ wardrobe}. Sadly, Lily never & . | alone | again | without | anymore \\
 &  & 3.37e-01 & 'm going to protect the\textbf{ castle} from the dragons," said & from | ," | with | !" | and \\
 &  & 3.59e-01 &  they needed for the tea\textbf{ party}. She brought out the & . | , | and | with | for \\
 &  & 3.62e-01 &  "I enjoy eating healthy\textbf{ snacks}." ĊĊThe moral & , | and | !" | like | too \\
 &  & 3.62e-01 &  and a kiss on the\textbf{ cheek}. From that day on & . | to | for | before | , \\
 &  & 3.68e-01 &  two of them enjoyed their\textbf{ juice}.<|endoftext|> & together | . | and | , | in \\
\end{longtable}
\end{landscape}

\section{Additional details}

\subsection{Compute resources}
\label{appsec:compute-resources}

We ran our experiments on two internal Slurm-based clusters equipped with nodes of 8 Nvidia H100 GPUs with 94GB of high-bandwidth memory per GPU or 8 Nvidia H200 GPUs with 141GB of high-bandwidth memory per GPU respectively.
The longest of our training runs takes approximately 12 hours on a node of 8 H200 GPUs to complete.
In total we spent the equivalent of around $\num{10000}$ H200 GPU hours to run our experiments.

\subsection{Software and libraries}
\label{appsec:software}
For the results obtained in this paper we build on free and open-source software.
We implemented our experiments in Python using JAX \citep[][Apache License 2.0]{bradbury_jax_2018}, Flax \citep[][Apache License 2.0]{heek_flax_2023}, the Deepmind JAX Ecosystem \citep[][Apache License 2.0]{babuschkin_deepmind_2020}, Grain \citep[][Apache License 2.0]{ritter_grain_2023}, HuggingFace dataset \citep[][Apache License 2.0]{lhoest_datasets_2021}, HuggingFace tokenizers \citep[][Apache License 2.0]{wolf_transformers_2020} and Scikit-learn \citep[][BSD 3-Clause License]{pedregosa_scikit-learn_2011}.
We utilized WandB \citep[][MIT license]{biewald_experiment_2020} to monitor the progress and results of experiments, and Plotly  \citep[][MIT license]{inc_collaborative_2015} for generating the plots.
We use uv for Python project dependency management \citep[][MIT License]{marsh_uv_2024}.

\end{document}